\documentclass{article}

\usepackage{iclr2026_conference,times}

\usepackage{amsmath,amsfonts,bm}

\def\eqref#1{equation~\ref{#1}}

\def\1{\bm{1}}

\DeclareMathAlphabet{\mathsfit}{\encodingdefault}{\sfdefault}{m}{sl}
\SetMathAlphabet{\mathsfit}{bold}{\encodingdefault}{\sfdefault}{bx}{n}

\usepackage[utf8]{inputenc}
\usepackage[T1]{fontenc}
\usepackage{hyperref}
\usepackage{url}
\usepackage{booktabs}
\usepackage{amsfonts}
\usepackage{amsmath}
\usepackage{nicefrac}
\usepackage{microtype}
\usepackage{xcolor}
\usepackage{colortbl}
\usepackage{graphicx}
\usepackage{float}
\usepackage{wrapfig}
\usepackage[skip=0.5\baselineskip]{caption}
\usepackage{bm}
\usepackage{multirow}
\usepackage{tabularx}
\usepackage{array}
\usepackage{enumitem}
\usepackage{threeparttable}
\usepackage{xspace}
\usepackage[normalem]{ulem}
\usepackage{algorithm}
\usepackage{algorithmicx}
\usepackage{algpseudocode}

\hypersetup{hidelinks}

\newcommand{\name}{PathNavigate\xspace}

\newcommand{\ie}{\emph{i.e.,}\xspace}

\newcommand{\best}[1]{{\boldmath$#1$}}
\newcommand{\runner}[1]{\underline{$#1$}}

\title{\name: A Training-Free Pathology Agent with Surprise-Guided Scan and Shared Slide Memory for Whole-Slide Image VQA}

\author{\normalfont Chunze Yang\textsuperscript{1},
Qidong Liu\textsuperscript{1\thanks{Corresponding authors: Qidong Liu and Chen Li}},
Wenjie Zhao\textsuperscript{1},
Yue Tang\textsuperscript{1},
Jiusong Ge\textsuperscript{1},
Di Zhang\textsuperscript{1},\\
\normalfont Jiashuai Liu\textsuperscript{1},
Lei Wu\textsuperscript{1},
Junbo Lu\textsuperscript{1},
Ni Zhang\textsuperscript{1},
Xian Wu\textsuperscript{2},
Zeyu Gao\textsuperscript{3},
Chen Li\textsuperscript{1}\\
\normalfont \textsuperscript{1} School of Comp. Science \& Technology, Xi'an Jiaotong University\\
\normalfont \textsuperscript{2} Tencent Jarvis Lab\\
\normalfont \textsuperscript{3} University of Cambridge\\
\normalfont\small \texttt{\{pureeeee,zhao\_wenjie,2206123832,JiusongGe,dzhang,}\\
\normalfont\small \texttt{ljs1007599414,2216113083,lujunbo\}@stu.xjtu.edu.cn}\\
\normalfont\small \texttt{\{liuqidong,nizhang,cli\}@xjtu.edu.cn}, \texttt{kevinxwu@tencent.com}\\
\normalfont\small \texttt{zg323@cam.ac.uk}
}

\iclrfinalcopy

\begin{document}
\raggedbottom

\maketitle
\fancyhead{}
\renewcommand{\headrulewidth}{0pt}

\begin{abstract}
Whole-slide image visual question answering (WSI-VQA) frames pathology as an extreme-context search problem: to answer a free-form clinical query, a system must first navigate a gigapixel slide under a strict inspection budget to locate sparse, high-resolution evidence.
Existing approaches largely fall into two paradigms: 
i) supervised pathology multimodal large language models (MLLMs) and agents can absorb localization and reasoning into learned modules, but they often couple navigation to task-specific supervision and retraining, limiting their practicality; 
ii) 
training-free pathology agents avoid this cost by keeping core models frozen, but often follow a question-first design, \ie constructing the initial candidate set mainly from query-conditioned relevance. This can miss decisive morphology that is not named in the question, and force heavier inference-time scaffolding. 
To address this challenge, we introduce \textbf{\name}, a training-free pathology agent built around a \textit{scan-search-readout} routine. Before question matching, \name scans the current slide at low magnification with a shared online memory module over frozen pathology features, producing a slide-specific surprise field that marks an abnormal-region pool. It then applies question-conditioned PLIP relevance only within this pool to select high-magnification search targets. Finally, it extracts local high-magnification evidence and answers with a frozen perceptor-adjudicator stack, using the same online memory as slide-level context. Experiments on WSI-VQA and SlideBench-BCNB show that the proposed scan-search-readout design improves answer accuracy and yields more interpretable evidence-selection trajectories with higher efficiency.
The code is available online\footnote{\url{https://github.com/Pureeeee/PathNavigate_}}.
\end{abstract}

\section{Introduction}
\label{sec:intro}

Whole-slide image (WSI) analysis has become a foundation of digital pathology: tissue sections can be digitized as gigapixel slides and used for diagnostic review, grading, and computational assessment \citep{niazi2019digital,campanella2019clinical}. Yet a single slide is far too large to inspect exhaustively. WSI visual question answering (WSI-VQA) makes this challenge more explicit: given a slide and a free-form clinical question, the system must decide where to look and then answer in language \citep{chen2024wsi}. More broadly, WSI-VQA is a special form of extreme-context visual search, where the model must allocate a tiny diagnostic inspection budget over a massive gigapixel visual field and reason from sparse, high-resolution evidence.

\begin{figure*}[!t]
\centering
\includegraphics[width=0.95\textwidth]{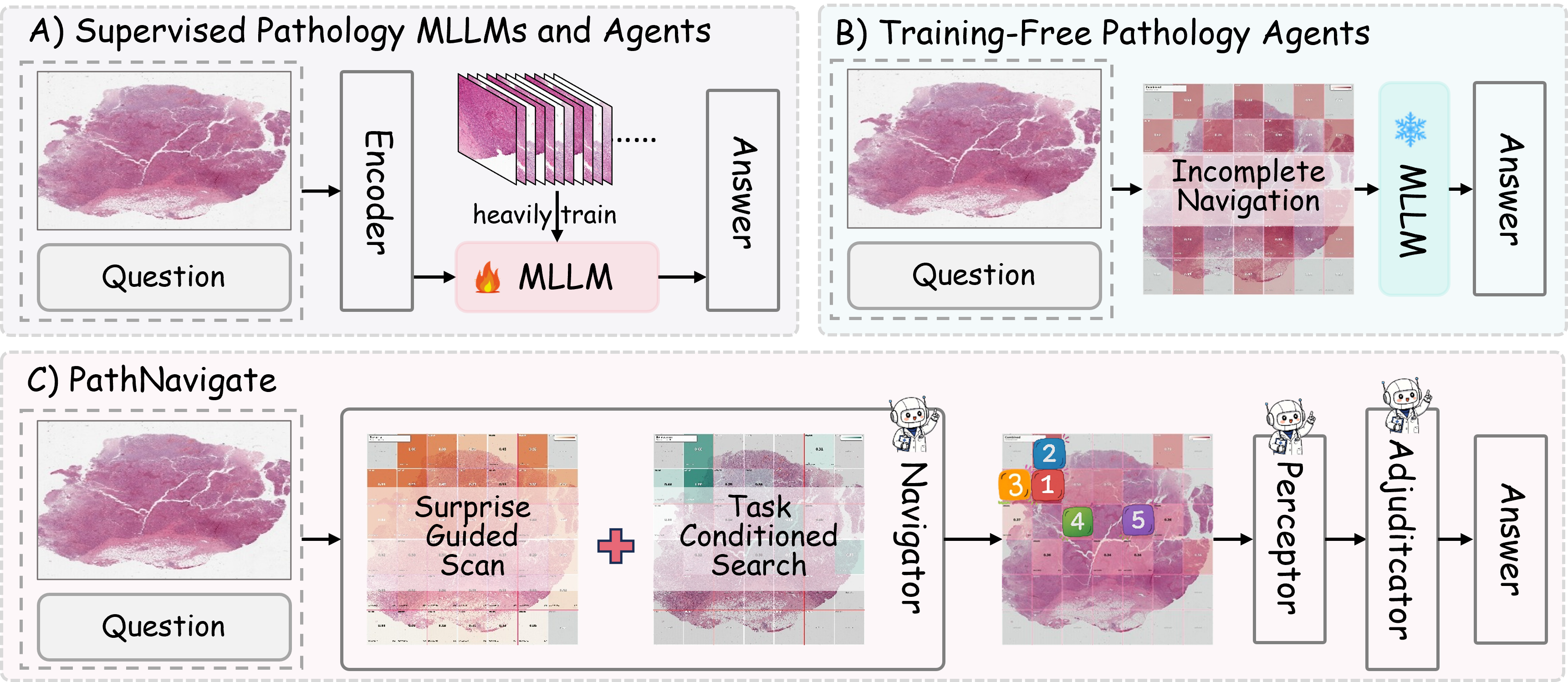}
\caption{The comparison between existing patterns and the proposed \textbf{Scan-Search-Readout} pattern for WSI-VQA.}
\label{fig:intro}
\end{figure*}

Recent progress in WSI-VQA generally follows two paradigms. (i) \textbf{Supervised Pathology Models and Agents}: slide-level multimodal large language models (MLLMs) such as SlideChat \citep{chen2025slidechat} and WSI-LLaVA \citep{liang2025wsi}, as well as trained agents such as CPathAgent \citep{sun2025cpathagent}, PathFinder \citep{pathfinder2025}, and SurvAgent \citep{survagent2025}, absorb evidence localization and reasoning into learned modules. 
Despite effectiveness, they couple search and reasoning to task-specific supervision, requiring retraining when the task or workflow changes. This limits their flexibility in deployment.
(ii) \textbf{Training-Free Pathology Agents}: by contrast, these agents keep the core MLLMs frozen and perform inference-time reasoning, with PathAgent as a representative example \citep{chen2025pathagent}. This line is increasingly popular because it preserves modularity and avoids task-specific retraining. 
Figure~\ref{fig:intro} contrasts these paradigms: supervised systems learn region localization inside the MLLM or agent stack, whereas training-free agents typically use reflection-driven navigation to revise their routes.


Since the training-free pattern relies primarily on reflected navigation for localization, the signal guiding the first pass over a gigapixel slide is crucial.
Recent works often adopt the \textbf{Question-First Design}, \ie the initial candidate set is constructed mainly from question-conditioned relevance rather than from a slide-specific content prior \citep{chen2025pathagent,huang2026actlike}.
However, this design leaves two problems. 
(i) \textbf{Incomplete Navigation}: pathology questions often underspecify the visual cue that determines the answer. For example, a question about histological type may require finding invasive ductal architecture, and a grading question may hinge on a small high-grade focus; neither cue is necessarily named in the question itself. A query-dependent scan can therefore retrieve generally relevant tumor regions while missing decisive morphology. 
(ii) \textbf{Heavier Inference Machinery}:
Due to weak scan signals, these agents rely more heavily on longer trajectories, reflection, caption caches, or external context \citep{chen2025pathagent,zhang2026pathrag,li2025tissuelab}, which can recover accuracy but increase latency, token consumption, and system complexity, undermining the practical appeal of training-free deployment.

To address the challenge mentioned above, we start from a pathology intuition: pathologists usually \underline{scan} a slide broadly to establish what is typical in the current case before \underline{searching} for the finding most relevant to the question. In digital pathology, broader scanning behavior is also more associated with diagnostic accuracy than zooming alone \citep{drew2021more}. This motivates a three-stage \textbf{Scan-Search-Readout} routine.
Before question matching, the agent should scan the current slide to build a slide-specific content prior, so diagnostically unusual regions are not discarded simply because they are not named by the question. During search, it should inject question relevance only within this content-aware candidate set, so the query refines rather than replaces slide-level exploration. During readout, it should reuse the same online state as slide-level memory for answering, rather than introducing a second store after navigation. Figure~\ref{fig:intro} visualizes the resulting shift: from directly retrieving question-matched regions to first scanning the current case, then searching within a compact candidate set, and finally answering with memory tied to the same scan.

In this paper, we instantiate this idea with the proposed \textbf{\name}, a training-free pathology agent for WSI-VQA. Our aim is to answer the design question above: how can a training-free agent establish a slide-specific search prior before question matching, while keeping answer-time slide context coupled to the same state that supports navigation? \name first scans the slide at low magnification and builds a compact candidate set from a slide-specific surprise prior over frozen pathology features using a small online memory module inspired by test-time memory methods \citep{behrouz2024titans,sun2020test}. It then applies question-conditioned PLIP relevance only within that set \citep{huang2023visual}, so task information refines the scan rather than replacing it. Finally, it extracts regional evidence and answers with slide-level context read out from the same online state used during scanning. At a broader level, \name provides a test-time framework for extreme-context visual search under tight inspection budgets.
Our contributions are three-fold:
\begin{enumerate}[leftmargin=16pt,itemsep=2pt]
\item We identify the limitation of question-first navigation in training-free WSI-VQA and propose a \textbf{Scan-Search-Readout} routine. The routine first builds a slide-specific content prior, then applies question relevance within that candidate set, and reuses the scan state for the answer.
\item We instantiate this routine as \name, a training-free pathology agent. \name uses a small online memory module over frozen pathology features to produce surprise-guided scan scores, applies PLIP only as within-pool reranking, and reads out the same online state as shared slide memory for answer aggregation.
\item We validate the design on WSI-VQA and SlideBench-BCNB. Overall performance comparison and ablations show that the scan-search-readout design improves answer accuracy and produces more interpretable evidence-selection trajectories without task-specific training.
\end{enumerate}


\section{Problem Setup}

\label{ssec:wsi_agent_framework}
Given a whole-slide image $\mathcal{W}$, a free-form question $q$, and frozen visual and language backbones, a training-free WSI-VQA system produces an answer $a$ after observing only a small subset of ROIs in the slide. Let $\mathcal{X}_{\ell}(\mathcal{W})=\{x_i^{\ell}\}_{i=1}^{N_\ell}$ and $\mathcal{X}_{h}(\mathcal{W})=\{x_j^{h}\}_{j=1}^{N_h}$ denote the low- and high-magnification tiles extracted from $\mathcal{W}$, and let $z=\phi(x)\in\mathbb{R}^{d}$ be the frozen pathology feature of a tile. The system may inspect only a bounded evidence set before producing the final answer, so the central problem is not only answering, but also how to allocate a small diagnostic visual budget over a massive gigapixel slide at test time.

Most training-free WSI-VQA pipelines can be described by four generic operations. 
(i) The slide is tiled at one or more magnifications and encoded by a frozen pathology model. 
(ii) A region-proposal step selects a compact region-of-interest (ROI) pool of candidate regions from the slide. 
(iii) The question prioritizes which candidate regions should receive the limited high-resolution evidence budget. 
(iv) A frozen pathology VLM, together with a frozen language model, aggregates regional evidence and slide-level context into the final answer.

\name keeps this same input-output setting and keeps all visual and language models frozen. Its difference lies in how the middle operations are instantiated. For operation (ii), \name proposes the ROI pool $\mathcal{R}$ from a slide-specific content prior built at $\ell$, rather than constructing candidates directly from question relevance. In Section~\ref{sec:method}, this pool is written as $\mathcal{R}(q)$ when the question changes only the routing budget and spatial spacing. For operation (iii), question relevance is used only to prioritize candidates within $\mathcal{R}$, producing the search targets $\mathcal{C}$. For operation (iv), local high-magnification evidence $\mathcal{E}$ is aggregated together with a readout of the same online state that produced the slide-specific content prior. In this terminology, our scan, search, and readout stages correspond to ROI proposal, question-based prioritization, and evidence aggregation, respectively.

\section{\name}
\label{sec:method}

\begin{figure*}[t]
\centering
\includegraphics[width=0.98\textwidth]{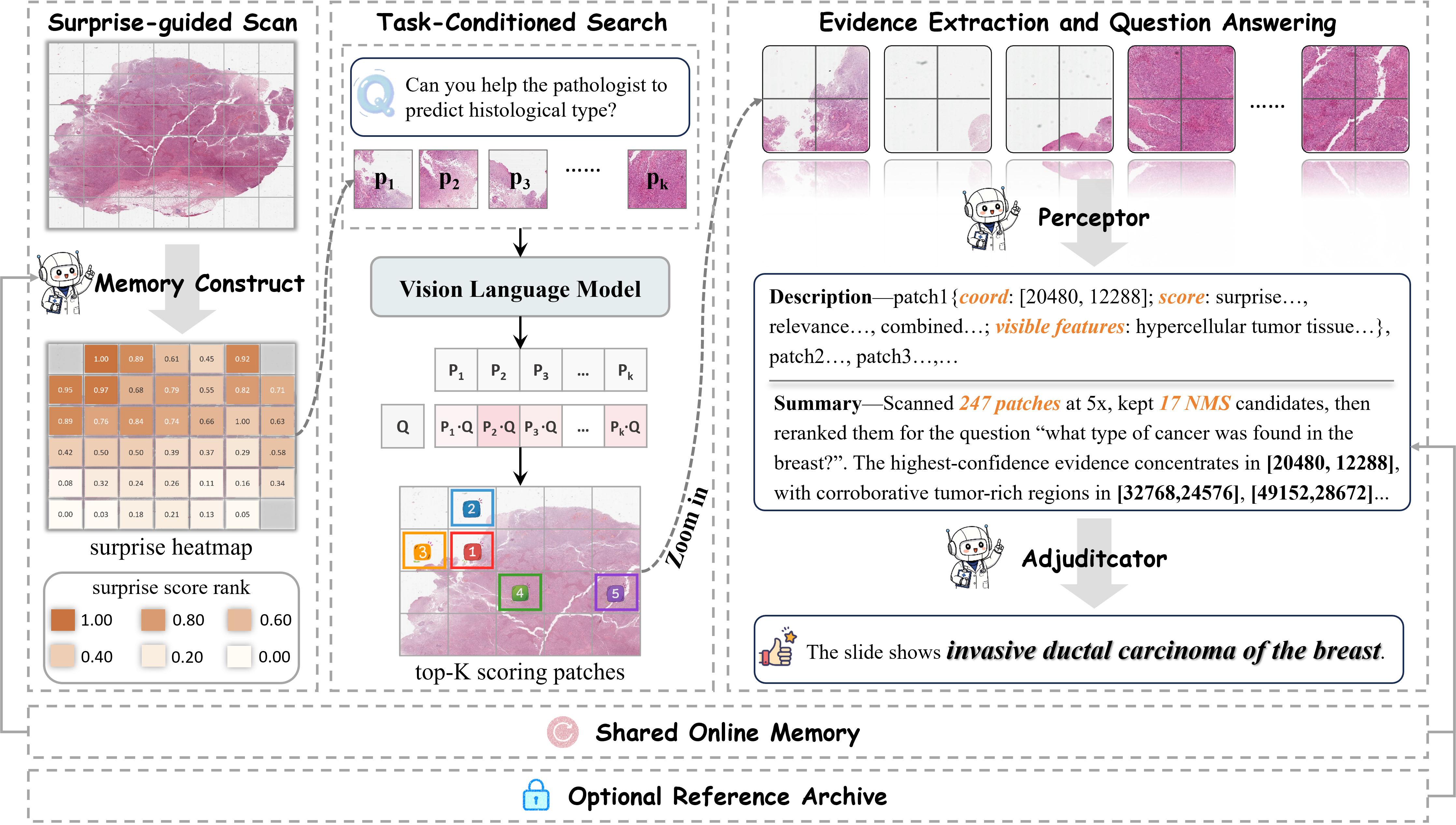}
\caption{Overview of \name as a scan-search-readout pathology agent.}
\label{fig:arch}
\end{figure*}

\vspace{1mm}
\noindent\textbf{Overview.}
Given a slide $\mathcal{W}$ and a question $q$, \name instantiates scan-search-readout with four modules: \emph{Shared Online Memory}, \emph{Surprise-Guided Scan}, \emph{Task-Conditioned Search}, and \emph{Evidence Readout and QA}. The pipeline builds a low-magnification surprise field, a compact ROI pool $\mathcal{R}(q)$, selected search targets $\mathcal{C}(q)$, and a final evidence set $\mathcal{E}(q)$ before returning answer $a$.
\textbf{Shared Online Memory} maintains a per-slide state over frozen CONCH features and supplies both scan-time surprise scores and compact slide context. \textbf{Surprise-Guided Scan} uses these scores to propose the low-magnification ROI pool; \textbf{Task-Conditioned Search} applies PLIP relevance only within that pool to select high-magnification targets; and \textbf{Evidence Readout and QA} converts selected patches into regional evidence for the final answer. Notably, the same scan state ties \emph{where to look} to \emph{what slide-level context is exposed}. We first define the memory, then describe the three operational stages.

\subsection{Shared Online Memory}
\label{ssec:mem}

Shared memory couples ROI proposal and answer-time context through one per-slide scan state. During scanning, tiles that would substantially change this state are treated as surprising; this process is denoted as \textbf{Memory Update}. 
At answer time, \textbf{Memory Readout} exposes a compact summary of the same surprise history and proposal process, so no separate learned store is introduced.

\vspace{1mm}
\noindent\textbf{Construction and Initialization.}
\label{sssec:online_state}
Let $x$ be a tile and let $\mathbf{z}=\phi(x)\in\mathbb{R}^{d}$ be its frozen CONCH feature, with $d=768$ \citep{lu2024conch}. The memory is a two-layer GELU reconstruction MLP,
\begin{equation}
\mathcal{M}_{\boldsymbol{\theta}}(\mathbf{z})=\mathbf{W}_2\,\mathrm{GELU}(\mathbf{W}_1\mathbf{z}+\mathbf{b}_1)+\mathbf{b}_2,
\qquad
\mathcal{M}_{\boldsymbol{\theta}}:\mathbb{R}^{d}\rightarrow\mathbb{R}^{d},
\label{eq:memory_mlp}
\end{equation}
where $\boldsymbol{\theta}=\{\mathbf{W}_1,\mathbf{b}_1,\mathbf{W}_2,\mathbf{b}_2\}$ denotes the online parameters.
The hidden width is $d=768$, and a fresh memory $\mathcal{M}_{\boldsymbol{\theta}_0}$ is instantiated for each slide, making surprise a within-slide signal rather than a comparison to a learned cohort. Initialization and optimization defaults are listed in Appendix~\ref{sec:appendix:routing}.

\vspace{1mm}
\noindent\textbf{Memory Update.}
Following test-time memory methods \citep{behrouz2024titans,sun2020test}, the memory adapts to the tile stream with a self-supervised reconstruction objective. A tile is surprising when reconstructing it would strongly change the current memory, so we use the pre-update gradient norm as surprise. For tile $x_i$ with feature $\mathbf{z}_i=\phi(x_i)$, the Huber reconstruction loss at state $\boldsymbol{\theta}_{i-1}$ is,
\begin{equation}
\begin{aligned}
\mathcal{L}_\delta(\boldsymbol{\theta}_{i-1};x_i)
&=\mathrm{Huber}_\delta\!\left(
\mathcal{M}_{\boldsymbol{\theta}_{i-1}}(\mathbf{z}_i),
\mathbf{z}_i
\right),
\end{aligned}
\label{eq:huber}
\end{equation}
where $\delta$ is the Huber transition parameter. The surprise of $x_i$ is the reconstruction-gradient norm before $x_i$ updates the memory,
\begin{equation}
\sigma_i=
\left\|\left.
\nabla_{\boldsymbol{\theta}}\mathcal{L}_\delta(\boldsymbol{\theta};x_i)
\right|_{\boldsymbol{\theta}=\boldsymbol{\theta}_{i-1}}\right\|_F.
\label{eq:surprise}
\end{equation}
Scoring before updating measures unexpectedness under earlier slide context. During the warm-up window $T_w$, every tile updates memory; after warm-up, we set $\tau_{\mathrm{sur}}=\mu_w+\lambda s_w$ from the first $T_w$ scores. Tiles with $\sigma_i>\tau_{\mathrm{sur}}$ trigger one clipped SGD step on Eq.~\eqref{eq:huber}; other tiles apply multiplicative decay $\boldsymbol{\theta}_i\leftarrow\alpha_f\boldsymbol{\theta}_{i-1}$.

\vspace{1mm}
\noindent\textbf{Memory Readout.}
\label{sssec:short}
At answer time, the adjudicator receives a short \emph{Navigation Summary} rather than raw MLP weights or a long trace. It reports the running surprise mean, standard deviation, high-surprise fraction, and candidate count; Appendix~\ref{sec:appendix:nav_prompt} gives the prompt.

\subsection{Surprise-Guided Scan}
\label{ssec:nav}

The scan stage turns a gigapixel slide into a compact set of regions for closer inspection. Its goal is to build a slide-specific content prior before text-image reranking: online memory highlights regions atypical for the current slide, so we use the \textbf{Surprise Score} as a proposal prior rather than a pathology label.
Here, the question acts only as a router for coverage and budget, not as a relevance scorer.
\textbf{ROI Selection} then turns surprise scores into target regions.

\vspace{1mm}
\noindent\textbf{Surprise Score.}
Let $\mathcal{X}_{\ell}(\mathcal{W})=\{x_i^{\ell}\}_{i=1}^{N_\ell}$ denote the low-magnification tiles of slide $\mathcal{W}$, with grid coordinates $\mathbf{p}_i$ and frozen CONCH features $\mathbf{z}_i=\phi(x_i^\ell)\in\mathbb{R}^{d}$. Applying the memory rule in Section~\ref{ssec:mem} gives each tile a surprise score: $\sigma_i=\sigma(x_i^{\ell})$ for $i=1,\dots,N_\ell$, 
where $\sigma(\cdot)$ is the pre-update gradient norm in Eq.~\eqref{eq:surprise}. Collecting these scores over all low-magnification tiles yields a slide-wide surprise field.

\vspace{1mm}
\noindent\textbf{ROI Selection.}
After collecting a slide-wide surprise field, we apply thresholding and distance-based non-maximum suppression (NMS) to the whole field to obtain a spatially diverse ROI pool:
\begin{equation}
\begin{aligned}
\mathcal{R}(q)
&=\operatorname{NMS}\!\left(
\{(\mathbf{p}_i,\sigma_i)\}_{i=1}^{N_\ell};\rho(q)
\right),
\end{aligned}
\label{eq:nms}
\end{equation}
where $\rho(q)$ is a minimum center distance in level-0 pixels rather than an IoU threshold. 
The operator sorts candidate centers by $\sigma_i$ and greedily keeps a center only if it is at least $\rho(q)$ away from all previously kept centers. $\mathcal{R}(q)$ is the resulting set of kept ROI centers, the handoff to Section~\ref{ssec:fusion}. 
As a result, morphology questions receive denser local coverage, while broader clinical questions receive larger spacing and a slightly larger first-pass search budget; Appendix~\ref{sec:appendix:routing} lists the reported values. 
This stage is intentionally question-light: the question influences spatial coverage and budget, but it does not yet rank candidate semantics.

\subsection{Task-Conditioned Search}
\label{ssec:fusion}

Once the scan has produced $\mathcal{R}(q)$, the system only needs to decide which regions in this slide-specific pool deserve high-magnification budget. Task-Conditioned Search therefore uses two complementary signals: surprise measures whether an ROI is atypical for the slide, while PLIP relevance measures whether it matches the question, \ie \textbf{Question-conditioned Relevance}. A transparent \textbf{Within-pool Fusion} combines them.
PLIP is applied at low magnification because this stage chooses which ROIs deserve follow-up.

\vspace{1mm}
\noindent\textbf{Question-conditioned Relevance.}
For each ROI center $r\in\mathcal{R}(q)$, we extract its corresponding low-magnification patch $x_r^\ell$ and compute a PLIP similarity score:
\begin{equation}
\begin{aligned}
r_{\mathrm{PLIP}}(r;q)
&=\cos\!\left(
\psi_{\mathrm{PLIP}}(q),
\phi_{\mathrm{PLIP}}(x_r^\ell)
\right),
\end{aligned}
\label{eq:relevance}
\end{equation}
where $\psi_{\mathrm{PLIP}}$ and $\phi_{\mathrm{PLIP}}$ are frozen PLIP text and image encoders \citep{huang2023visual}.

\vspace{1mm}
\noindent\textbf{Within-pool Fusion.}
Let $\sigma(r)$ denote the scan-time surprise score associated with ROI center $r$. Surprise and PLIP relevance have different numeric scales, and their ranges vary from slide to slide. Because this stage only needs the ordering inside the current ROI pool, we normalize each score function $s$ within $\mathcal{R}(q)$ to obtain a pool-relative score,
\begin{equation}
\hat s(r)=\frac{s(r)-m_s}{M_s-m_s+\varepsilon},
\label{eq:norm}
\end{equation}
where $s\in\{\sigma, r_{\mathrm{PLIP}}(\cdot;q)\}$, $m_s=\min_{r'\in\mathcal{R}(q)}s(r')$, and $M_s=\max_{r'\in\mathcal{R}(q)}s(r')$. The small $\varepsilon$ prevents division by zero when all ROI scores are tied. This produces $\hat\sigma(r)$ for slide-specific surprise and $\hat r_{\mathrm{PLIP}}(r;q)$ for question relevance. We then combine the two normalized scores as:
\begin{equation}
\begin{aligned}
f_\alpha(r;q)
&=\alpha\,\hat r_{\mathrm{PLIP}}(r;q)
+(1-\alpha)\,\hat\sigma(r),
\end{aligned}
\label{eq:fusion}
\end{equation}
where $\alpha\in[0,1]$ controls the balance between task relevance and slide-specific surprise. 
We use $\alpha=0.5$ as a fixed symmetric default and ablate the endpoints and sweep values in Section~\ref{ssec:exp:components}; the full numeric sweep is reported in Table~\ref{tab:appendix:alpha_numeric}. The final search targets are:
\begin{equation}
\mathcal{C}(q)=\operatorname{Top}K\!\bigl(\mathcal{R}(q),f_\alpha(\cdot;q);K_{\mathrm{search}}(q)\bigr),
\label{eq:search_select}
\end{equation}
where $K_{\mathrm{search}}(q)$ is a small question-aware first-pass budget. The same lightweight router that sets $\rho(q)$ allocates denser local follow-up to morphology questions and broader coverage to clinical questions; Appendix~\ref{sec:appendix:routing} lists the values. The role split remains asymmetric: surprise builds the pool, and PLIP only reranks within that pool.

\subsection{Evidence Readout and QA}
\label{ssec:qa}

Given search targets $\mathcal{C}(q)$ and the slide-level readout, the final stage collects answerable evidence: \textbf{Evidence Extraction} refines each ROI at high magnification, \textbf{Reference Archive} optionally retrieves context, and \textbf{Perceptor and Adjudicator} produce the answer.

\vspace{1mm}
\noindent\textbf{Evidence Extraction.}
For each selected low-magnification target $r\in\mathcal{C}(q)$, we collect its local high-magnification neighborhood $\mathcal{P}_{h}(r)$. When high-magnification features are available, we instantiate a fresh local memory for each ROI, so the stage ranks micro-features inside the selected region without inheriting or overwriting the global slide state. It applies the same surprise rule inside $\mathcal{P}_{h}(r)$ with a shorter ROI-size-adaptive warm-up schedule and keeps the top-$T$ patches,
\begin{equation}
\mathcal{E}(q)=\bigcup_{r\in\mathcal{C}(q)}\operatorname{Top}T\!\bigl(\mathcal{P}_{h}(r),\sigma_{h}(\cdot\,;r)\bigr),
\label{eq:evidence_select}
\end{equation}
where $\sigma_{h}(\cdot\,;r)$ is the local surprise score inside ROI $r$. Here, $T$ is a small per-ROI evidence budget, and the total number of perceptor calls is capped by a separate global budget; Appendix~\ref{sec:appendix:routing} lists the launch values. If high-magnification features are unavailable or disabled, the low-magnification target serves as fallback evidence.

\vspace{1mm}
\noindent\textbf{Reference Archive.}
\label{sssec:long}
The optional archive aids final adjudication, not navigation. Before adjudication, we may retrieve $k$ reference cases by cosine similarity between mean-pooled low-magnification CONCH slide features and append their short summaries as \emph{Reference Context}. The reported main results use $k=3$ unless an ablation disables the archive.

\vspace{1mm}
\noindent\textbf{Perceptor and Adjudicator.}
\label{sssec:consumers}
We use the frozen pathology VLM as \emph{perceptor} to turn patches into evidence strings, and the frozen language model as \emph{adjudicator} to produce the final answer from regional evidence, the Navigation Summary, and optional Reference Context. Concrete backbones, serving stack, and adjudication settings are reported in Section~\ref{ssec:exp:setup} and Appendix Table~\ref{tab:appendix:settings}.

\vspace{1mm}
\noindent\textbf{Inference Summary.}
At inference time, \name scans at low magnification to build $\mathcal{R}(q)$, reranks with PLIP to obtain $\mathcal{C}(q)$, reads high-magnification evidence $\mathcal{E}(q)$, and answers from regional evidence plus the shared memory readout. Appendix~\ref{sec:appendix:method_algorithm} provides pseudocode.

\section{Experiments}
\label{sec:exp}

\subsection{Setup}
\label{ssec:exp:setup}

\noindent\textbf{Benchmarks.} We evaluate on WSI-VQA~\citep{chen2024wsi}, SlideBench-BCNB~\citep{chen2025slidechat}, and PathMMU~\citep{sun2024pathmmu}. WSI-VQA and BCNB test whole-slide question answering, while PathMMU is used only as a patch-level transfer check. Dataset statistics, splits, and protocol details are deferred to Appendix~\ref{sec:appendix:exp_details}.

\vspace{1mm}
\noindent\textbf{Baselines and protocol.} We organize baselines into three groups aligned with the abstract taxonomy. \textbf{Pretrained MLLM Models} include GPT-4o~\citep{openai2024gpt4o}, GPT-5.4~\citep{openai2026models}, Gemini-3.1-Pro-preview~\citep{google2026gemini31pro}, and Qwen3.5-4B/9B~\citep{qwen2026qwen35}, evaluated as off-the-shelf single-thumbnail systems. \textbf{Supervised Pathology Models} include the WSI-VQA model~\citep{chen2024wsi}, SlideChat~\citep{chen2025slidechat}, WSI-LLaVA~\citep{liang2025wsi}, MedDr~\citep{he2024gsco}, TITAN~\citep{TITAN}, Quilt-LLaVA~\citep{quiltllava2024}, LLaVA-Med~\citep{li2023llava} where available, MedGemma-1.5-4B~\citep{sellergren2025medgemma}, and Patho-R1 only~\citep{zhang2026pathor1}; these use pathology-specific training, medical/pathology adaptation, or WSI foundation-model pretraining, but not a training-free WSI navigation loop. \textbf{Training-Free Pathology Agents} include PathAgent~\citep{chen2025pathagent} and \name. Appendix~\ref{sec:appendix:baseline_settings} gives baseline settings, model identifiers, and evaluation joins.

\vspace{1mm}
\noindent\textbf{Implementation Details.} \name uses Patho-R1-7B~\citep{zhang2026pathor1} as frozen perceptor, Qwen3.5-4B~\citep{qwen2026qwen35} as adjudicator, and CONCH~v1.5~\citep{lu2024conch} for patch features; Qwen endpoints use vLLM~\citep{kwon2023efficient} where required. Appendix~\ref{sec:appendix:exp_details} records hardware/software, and Appendix~\ref{sec:appendix:routing} records routing budgets and launch caps.

\vspace{1mm}
\noindent\textbf{Evaluation Metrics}
For WSI-VQA, we report Total accuracy, MCQ accuracy, open-ended accuracy, BLEU-1, and ROUGE-L as percentages in the main table, with BLEU-4 in Appendix~\ref{sec:appendix:tables}. For BCNB and PathMMU, we use letter-exact accuracy; BCNB Overall follows the official weighted aggregation over seven diagnostic sub-tasks. Efficiency reports latency, token count, and offline footprint, where lower values are preferred.
\subsection{Main results}
\label{ssec:exp:main}

\noindent\textbf{Overall comparison on SlideBench-BCNB (Table~\ref{tab:bcnb}).}
BCNB is fully multiple-choice and reports diagnostic subscores, making it useful for comparing method families. \name improves Overall by building a slide-specific candidate pool before question matching and then answering from high-magnification evidence. The strongest gains over the frozen perceptor appear on Tumor, ER, PR, and the merged HER2 score, where locating invasive regions and preserving regional morphology directly improves the answer context. Across models, the merged HER2 column is still compressed relative to Tumor and ER because the expression-level part is less directly recoverable from H\&E morphology. Thus \name is best on Overall, ER, HER2, and Molecular Subtype, runner-up on Tumor and PR, and competitive but not dominant on Grading. This pattern is consistent with the design goal: the scan-search-readout loop improves evidence localization, while tasks requiring information outside visible morphology remain harder even after better routing.

\begin{table}[!t]
\centering
\caption{Primary SlideBench-BCNB results with full sub-task breakdown. The boldface refers to the highest score and the underline indicates the next best result among all methods.}
\label{tab:bcnb}
\scriptsize
\setlength{\tabcolsep}{2.4pt}
\renewcommand{\arraystretch}{1.2}
\resizebox{\linewidth}{!}{%
\begin{tabular}{llccccccc}
\toprule
Method & Input & Overall & Tumor & ER & PR & HER2 & Grading & Mol.Sub \\
\midrule
\multicolumn{9}{l}{\textit{Pretrained MLLM Models}} \\
GPT-4o~\citep{openai2024gpt4o}                            & Slide-T & $43.78$ & $56.84$ & $49.96$ & $49.42$ & $42.13$ & $38.46$ & $26.84$ \\
GPT-5.4~\citep{openai2026models}                           & Slide-T & $45.85$ & $60.42$ & $50.42$ & $49.86$ & $44.51$ & $42.36$ & $28.43$ \\
Gemini-3.1-Pro-preview~\citep{google2026gemini31pro}       & Slide-T & $44.91$ & $58.46$ & $50.18$ & $49.62$ & $43.58$ & $40.85$ & $27.62$ \\
Qwen3.5-4B~\citep{qwen2026qwen35}                         & Slide-T & $39.48$ & $49.43$ & $33.36$ & $27.32$ & \runner{48.02} & $52.70$ & $19.19$ \\
Qwen3.5-9B~\citep{qwen2026qwen35}                         & Slide-T & $41.99$ & $53.84$ & $36.42$ & $30.18$ & $47.53$ & \best{59.15} & $21.43$ \\
\midrule
\multicolumn{9}{l}{\textit{Supervised Pathology Models}} \\
SlideChat~\citep{chen2025slidechat}                       & WSI     & $54.07$ & \best{89.90} & \runner{72.98} & \best{72.98} & $47.53$ & $23.11$ & $20.58$ \\
WSI-LLaVA~\citep{liang2025wsi}                            & WSI     & $52.68$ & $85.32$ & $60.81$ & $60.81$ & $42.78$ & $46.28$ & $29.20$ \\
MedDr~\citep{he2024gsco}                                 & Slide-T & $35.48$ & $59.60$ & $43.80$ & $39.20$ & $27.40$ & $29.40$ & $20.80$ \\
TITAN~\citep{TITAN}                                      & Slide   & $44.39$ & $85.16$ & $42.44$ & $41.68$ & $36.86$ & $45.35$ & $22.49$ \\
Quilt-LLaVA~\citep{quiltllava2024}                        & Patch   & $34.11$ & $33.46$ & $49.43$ & $44.05$ & $25.85$ & $34.99$ & $25.24$ \\
MedGemma-1.5-4B~\citep{sellergren2025medgemma}            & Slide-T & $50.38$ & $79.68$ & $52.36$ & $60.35$ & $47.85$ & $42.33$ & $21.27$ \\
Patho-R1 only~\citep{zhang2026pathor1}                    & Patch   & $37.26$ & $42.16$ & $54.63$ & $50.38$ & $25.62$ & $35.53$ & $26.65$ \\
\midrule
\multicolumn{9}{l}{\textit{Training-Free Pathology Agents}} \\
PathAgent~\citep{chen2025pathagent}                       & WSI     & \runner{55.72} & $87.52$ & $63.69$ & $63.69$ & $44.52$ & \runner{55.95} & \runner{30.21} \\
\rowcolor{gray!10}
\textbf{\name (ours)}                                     & WSI     & \best{59.42} & \runner{88.28} & \best{76.09} & \runner{69.00} & \best{48.40} & $52.58$ & \best{32.33} \\
\bottomrule
\end{tabular}
}
\end{table}

\begin{table}[!t]
\centering
\caption{Compact WSI-VQA results. All numeric scores are percentages. TITAN reports only MCQ because this zero-shot setting can only be mapped to answer options; boldface/underline mark best/runner-up.}
\label{tab:wsivqa}
\fontsize{9.5pt}{9pt}\selectfont
\setlength{\tabcolsep}{4.2pt}
\renewcommand{\arraystretch}{1.1}
\begin{tabularx}{\linewidth}{@{}Xlccccc@{}}
\toprule
Method & Input & Total & MCQ & Open & BLEU-1 & ROUGE-L \\
\midrule
\multicolumn{7}{l}{\textit{Pretrained MLLM Models}} \\
GPT-4o~\citep{openai2024gpt4o}                 & Slide-T & $27.57$ & $32.50$ & $22.00$ & $8.53$ & $23.62$ \\
GPT-5.4~\citep{openai2026models}                & Slide-T & $35.74$ & $42.18$ & $28.46$ & $14.27$ & $28.86$ \\
Gemini-3.1-Pro-preview~\citep{google2026gemini31pro} & Slide-T & $34.41$ & $40.42$ & $27.62$ & $12.84$ & $27.56$ \\
Qwen3.5-4B~\citep{qwen2026qwen35}              & Slide-T & $40.26$ & $40.26$ & $40.26$ & $9.21$ & $24.13$ \\
Qwen3.5-9B~\citep{qwen2026qwen35}              & Slide-T & $38.50$ & $49.23$ & $26.38$ & \runner{57.31} & \runner{48.04} \\
\midrule
\multicolumn{7}{l}{\textit{Supervised Pathology Models}} \\
WSI-VQA model~\citep{chen2024wsi}              & WSI     & $42.18$ & $46.90$ & $36.84$ & $10.83$ & $31.79$ \\
MedDr~\citep{he2024gsco}                       & Slide-T & \runner{45.18} & $43.69$ & \runner{46.86} & $36.82$ & $45.24$ \\
TITAN~\citep{TITAN}                            & Slide   & ---     & $34.87$ & ---     & ---     & ---     \\
Quilt-LLaVA~\citep{quiltllava2024}             & Patch   & $30.07$ & $29.23$ & $31.01$ & $0.24$ & $25.13$ \\
WSI-LLaVA~\citep{liang2025wsi}                 & Patch   & $33.20$ & $42.82$ & $22.32$ & $5.73$ & $31.24$ \\
MedGemma-1.5-4B~\citep{sellergren2025medgemma} & Slide-T & $42.99$ & \best{53.08} & $31.59$ & $18.04$ & $34.93$ \\
\midrule
\multicolumn{7}{l}{\textit{Training-Free Pathology Agents}} \\
PathAgent~\citep{chen2025pathagent}            & WSI     & $33.88$ & $51.54$ & $13.91$ & $42.73$ & $33.54$ \\
\rowcolor{gray!10}
\textbf{\name (ours)}                         & WSI     & \best{56.34} & \runner{52.21} & \best{61.00} & \best{60.92} & \best{63.53} \\
\bottomrule
\end{tabularx}
\vspace{-0.6em}
\end{table}

\noindent\textbf{Overall comparison on WSI-VQA (Table~\ref{tab:wsivqa}).}
WSI-VQA complements BCNB with multiple-choice and open-ended answers. Thumbnail and patch-only baselines can solve some MCQs but lack reliable evidence recovery for free-form responses. Under the same Patho-R1-7B/Qwen3.5-4B stack, PathAgent remains competitive on MCQ yet drops on Open accuracy and ROUGE-L, indicating that \name's gains come from scan-derived context and high-magnification readout rather than a decoder swap. The contrast between MCQ and Open is informative: MedGemma and PathAgent can stay close on closed-set choices, but neither maintains the evidence chain needed to produce a correct free-form answer. \name's Open score and text-overlap gains therefore reflect not only answer selection but also more complete regional evidence coverage.

\noindent\textbf{Cross-benchmark reading.}
Together, Tables~\ref{tab:bcnb} and~\ref{tab:wsivqa} show that the main bottleneck is evidence routing rather than raw model scale alone. Strong pretrained or supervised baselines remain competitive on constrained sub-tasks, but they either compress the slide into a small view or rely on learned pathology priors. \name is strongest when the answer requires finding and verbalizing the right region under a limited budget, which explains the simultaneous gains in BCNB Overall, WSI-VQA Open accuracy, and text-overlap metrics. The residual columns also delimit the claim: when a label depends on IHC-like information or grading cues that are diffuse at the selected scale, better navigation helps but cannot fully substitute for missing modality evidence.

Appendix Table~\ref{tab:appendix:settings} gives settings; Appendix Figures~\ref{fig:appendix:wsivqa_agent_compare} and~\ref{fig:appendix:bcnb_taskwise} summarize controlled comparisons; Appendix~\ref{sec:appendix:pathagent} documents the PathAgent rerun gap. \noindent\textbf{PathMMU transfer.} Because PathMMU is image-level, Appendix Table~\ref{tab:appendix:pathmmu_full} reports it only as patch-level validation ($53.60$ letter-exact accuracy).

\subsection{Core component analysis}
\label{ssec:exp:components}

The ablations map to scan-search-readout: \textit{w/o} PLIP removes question-aware reranking, PLIP-only removes surprise, Random ROI removes structured proposal, and archive/Navigation Summary variants test answer-time context. Table~\ref{tab:ablation_wsivqa} supports a division of labor: surprise proposes, PLIP refines, high-magnification evidence grounds the answer, and compact context stabilizes adjudication. The largest drop comes from replacing the proposal mechanism with random ROIs ($-6.84$ Total), showing that the first-pass pool is the primary failure point; removing PLIP costs less but still hurts MCQ, where task wording helps discriminate among plausible regions. Archive and Navigation Summary ablations are smaller but consistent, suggesting that answer-time context mainly stabilizes borderline cases rather than creating the evidence signal. Additional BCNB, architecture, high-magnification readout (Table~\ref{tab:appendix:highmag_readout}), negative-result, and sensitivity details are deferred to Appendix~\ref{sec:appendix:sensitivity_figures}.

\begin{table}[H]
\centering
\caption{Structural WSI-VQA ablations.}
\label{tab:ablation_wsivqa}
\small
\setlength{\tabcolsep}{7pt}
\renewcommand{\arraystretch}{1.08}
\begin{tabular*}{0.72\linewidth}{@{\extracolsep{\fill}}lccc}
\toprule
Variant & Total & MCQ & Open \\
\midrule
\rowcolor{gray!10}
\name & \best{56.34} & \best{52.21} & \best{61.00} \\
\textit{w/o} PLIP & $53.42$ & $47.20$ & \runner{60.46} \\
PLIP-only & $52.20$ & $46.90$ & $58.20$ \\
\textit{w} Random ROI & $49.50$ & $45.18$ & $54.38$ \\
\textit{w/o} archive & $53.84$ & $49.50$ & $58.74$ \\
\textit{w/o} Nav. Sum. & \runner{54.82} & \runner{50.62} & $59.56$ \\
\bottomrule
\end{tabular*}
\vspace{-0.4em}
\end{table}

\subsection{System-level efficiency}
\label{ssec:exp:efficiency}
\begin{figure}[t]
\centering
\includegraphics[width=0.50\linewidth]{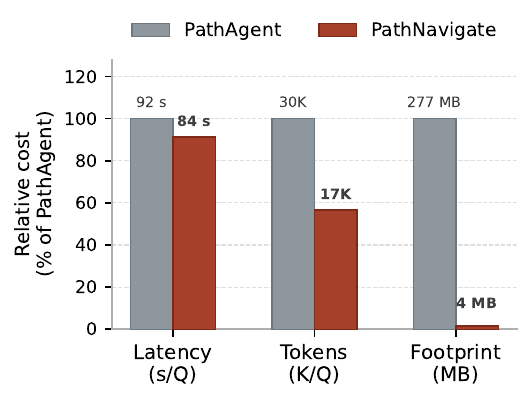}
\caption{Relative WSI-VQA cost for training-free agents.}
\label{fig:efficiency}
\end{figure}

We measure per-question cost for \name and PathAgent on the same hardware, WSI-VQA set, and evaluation pipeline. Figure~\ref{fig:efficiency} normalizes PathAgent to $100\%$: \name keeps latency similar while reducing prompt tokens and footprint (on-disk auxiliary artefacts, not model weights). Fewer prompt tokens reduce evidence passed into the adjudicator, lowering context pressure without removing high-magnification readout. The claim is scoped to storage and token budget against training-free agents, not peak GPU memory or parity with trained slide MLLMs; Appendix~\ref{sec:appendix:efficiency_plots} gives the breakdown.

\subsection{Qualitative case studies}
\label{ssec:exp:qualitative}

\begin{figure}[H]
\centering
\includegraphics[width=0.98\linewidth]{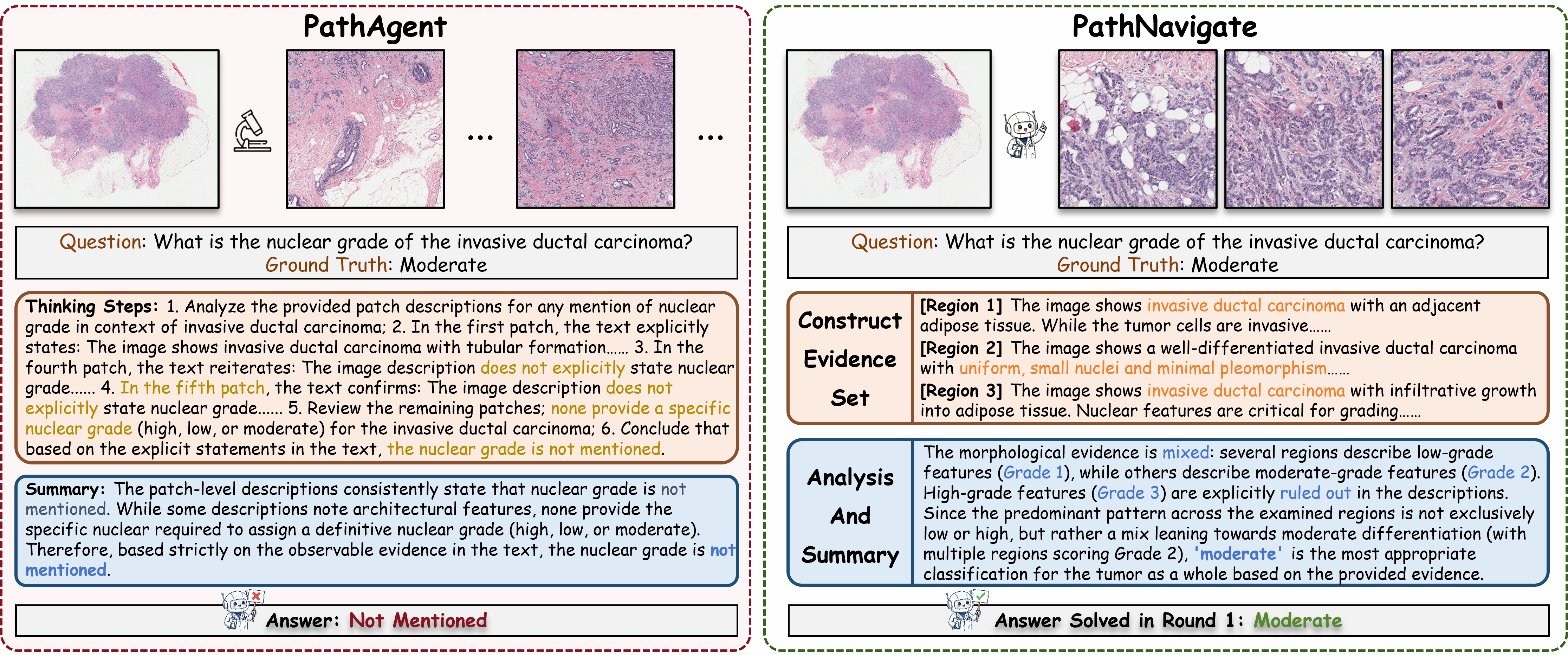}
\caption{Diagnostic pathway comparison. PathAgent follows weak tissue and predicts squamous cell carcinoma; \name uses a slide-wide prior and high-magnification evidence to predict infiltrating ductal carcinoma.}
\label{fig:pathagent_case_compare}
\end{figure}

Figure~\ref{fig:pathagent_case_compare} compares the routes for the same slide and question: PathAgent follows weak tissue, while \name reaches the ductal carcinoma label with a slide-wide prior and high-magnification evidence. The case shows adjudication must anchor to diagnostic tissue before answering. Appendix~\ref{sec:appendix:workflow} gives the full workflow and intermediate evidence trace.

\section{Related Work}
\label{sec:related_work}

\noindent\textbf{Supervised Pathology MLLMs and Agents.}
Pathology VQA requires gigapixel navigation with morphological and diagnostic reasoning. Earlier datasets and benchmarks, including PathVQA~\citep{he2020pathvqa}, Quilt-llava~\citep{seyfioglu2024quilt}, and WSICaption~\citep{chen2024wsicaption}, advanced medical VQA but only partially stress slide-level evidence search. Recent supervised pathology MLLMs such as SlideChat~\citep{chen2025slidechat} and WSI-LLaVA~\citep{liang2025wsi}, and trained agents such as CPathAgent~\citep{sun2025cpathagent}, PathFinder~\citep{pathfinder2025}, and SurvAgent~\citep{survagent2025}, absorb localization and reasoning into learned modules but depend on task-specific training or medical image-text data. These methods are useful upper references for pathology-specific adaptation, but they answer a different question from ours: how much slide reasoning can be learned with supervision, rather than how a frozen system should allocate inspection at test time. We therefore test navigation without task-specific retraining in this setting.

\noindent\textbf{Training-Free Pathology Agents.}
To avoid task-specific retraining~\citep{zhuang2026libra,sun2025cpath}, training-free agents formulate WSI-VQA as evidence gathering with frozen models. Existing systems often depend on engineered tools, caption caches, or external knowledge bases~\citep{wang2023chatcad,yu2025ypathrag,schmidgall2024agentclinic,tang2024medagents}, and many follow a question-first retrieval prior~\citep{karim2025multimodal,chen2025pathagent}. This can work when query terms align with visible tissue, but it makes early search inherit the user's wording and can miss morphology that is diagnostic but not named. \name's contribution is therefore not a larger tool inventory or retrieval database, but a change in the order of scan, search, and readout. \name instead scans first, builds the pool from intrinsic morphology, and applies the question only within the pool.

\section{Conclusion}

We presented \name, a training-free WSI-VQA agent built around scan-search-readout. Shared online memory creates a slide-specific surprise prior before question matching and ties navigation to answer-time evidence. Experiments on WSI-VQA and SlideBench-BCNB improve accuracy and evidence selection across pretrained MLLMs, supervised pathology models, and training-free agents, while reducing storage and token budget relative to the training-free PathAgent baseline. More broadly, the results suggest that test-time navigation is a first-class design variable even when all perception and language backbones remain frozen. H\&E-weak labels motivate richer modality-aware evidence for future training-free agents.

\clearpage
\bibliographystyle{iclr2026_conference}
\bibliography{main}

\clearpage
\appendix
\phantomsection
\begin{center}
\rule{0.94\linewidth}{1.2pt}

\vspace{1.6em}
{\Large\bfseries \name: A Training-Free Pathology Agent with Surprise-Guided Scan\\
and Shared Slide Memory for Whole-Slide Image VQA}\par
\vspace{0.25em}
{\large\bfseries\itshape Supplementary Material}\par
\vspace{1.7em}
\rule{0.94\linewidth}{0.4pt}
\end{center}

\vspace{1.5em}
\noindent{\Large\bfseries Table of Contents}
\vspace{0.2em}
\hrule
\vspace{0.9em}

\begingroup
\small
\makeatletter
\newcommand{\appdirsection}[3]{%
	\par\addvspace{0.95em}%
	\noindent\makebox[2.4em][l]{\bfseries #1}%
	{\bfseries #2}\nobreak\hfill{\bfseries \pageref{#3}}\par}
\newcommand{\appdirsubsection}[3]{%
	\@dottedtocline{1}{2.4em}{3.4em}{#1\quad #2}{\pageref{#3}}}
\makeatother
\appdirsection{A}{Experimental Settings and Reproduction Protocols}{sec:appendix:experimental_settings}
\appdirsubsection{A.1}{Experimental Setting Details}{sec:appendix:exp_details}
\appdirsubsection{A.2}{Baseline Settings, Model Names, and API Scope}{sec:appendix:baseline_settings}
\appdirsubsection{A.3}{PathAgent Reproduction Protocol}{sec:appendix:pathagent}

\appdirsection{B}{Method Details}{sec:appendix:method_details}
\appdirsubsection{B.1}{Question-aware Routing and Budget Allocation}{sec:appendix:routing}
\appdirsubsection{B.2}{Navigation Summary Prompt Example}{sec:appendix:nav_prompt}
\appdirsubsection{B.3}{Inference Pseudocode}{sec:appendix:method_algorithm}

\appdirsection{C}{Experimental Results and Additional Analyses}{sec:appendix:experimental_results}
\appdirsubsection{C.1}{Full Result Tables}{sec:appendix:tables}
\appdirsubsection{C.2}{Supplementary Comparison Plots}{sec:appendix:comparison_plots}
\appdirsubsection{C.3}{Workflow Gallery Template}{sec:appendix:workflow}
\appdirsubsection{C.4}{Sensitivity and Component Figures}{sec:appendix:sensitivity_figures}
\appdirsubsection{C.5}{Efficiency Visualization}{sec:appendix:efficiency_plots}
\appdirsubsection{C.6}{Negative Findings}{sec:appendix:negative}
\endgroup

\vfill
\hrule

\clearpage

\section{Experimental Settings and Reproduction Protocols}
\label{sec:appendix:experimental_settings}

This section collects the material needed to reproduce the experimental stack: benchmark definitions, baseline categories, implementation details, metric definitions, model/API scope, and the controlled PathAgent protocol.

\subsection{Experimental Setting Details}
\label{sec:appendix:exp_details}

This subsection expands the shortened experimental-setting paragraphs in the main paper, following the Dataset--Baselines--Implementation--Metrics organization used by the reference paper.

\noindent\textbf{Benchmarks.}
WSI-VQA~\citep{chen2024wsi} contains $735$ TCGA-BRCA questions, including $390$ multiple-choice and $345$ open-ended questions. We use the same test records and answer join for \name, the PathAgent row, and the reproduced baselines. SlideBench-BCNB~\citep{chen2025slidechat} contains $7{,}274$ BCNB multiple-choice questions over Tumor, ER, PR, HER2 status, HER2 expression, Grading, and Molecular Subtype; compact BCNB tables follow the main table and merge HER2 status with HER2 expression by averaging the two scores, while Overall keeps the official seven-task weighting. All local and reproduced rows use the full test set and the SlideChat letter-exact protocol. PathMMU~\citep{sun2024pathmmu} is image-level rather than WSI-level, so it is reported only as a patch-level transfer validation for the frozen perception and answering modules; the complete transfer table is collected in Table~\ref{tab:appendix:pathmmu_full}.

\vspace{1mm}
\noindent\textbf{Baseline categories.}
\begin{itemize}[leftmargin=16pt,itemsep=2pt]
\item \textbf{Pretrained MLLM Models.} GPT-4o, GPT-5.4, Gemini-3.1-Pro-preview, and Qwen3.5-4B/9B are treated as off-the-shelf MLLMs. They use one low-magnification slide thumbnail in the main comparison and do not add pathology-specific WSI navigation.
\item \textbf{Supervised Pathology Models.} This group covers the WSI-VQA model, SlideChat, WSI-LLaVA, MedDr, TITAN, Quilt-LLaVA, LLaVA-Med where available, MedGemma-1.5-4B, and Patho-R1 only. These rows use pathology-specific training, medical/pathology adaptation, or WSI foundation-model pretraining, but they are not training-free pathology agents in our comparison.
\item \textbf{Training-Free Pathology Agents.} PathAgent and \name use frozen core models with explicit WSI evidence-gathering loops. Direct agent-to-agent claims use the controlled PathAgent protocol rather than the original paper-reported MCQ-only number.
\end{itemize}

\vspace{1mm}
\noindent\textbf{Implementation stack.}
\name uses Patho-R1-7B~\citep{zhang2026pathor1} as the frozen perceptor, Qwen3.5-4B~\citep{qwen2026qwen35} as the frozen adjudicator, CONCH~v1.5~\citep{lu2024conch} for patch features, and vLLM~\citep{kwon2023efficient} for Qwen endpoint serving where required. Local runs were executed on a dual Intel Xeon Platinum 8352V CPU host with NVIDIA RTX 4090 24GB GPUs. The software stack is Python 3.9.25, PyTorch 2.7.0+cu126, CUDA 12.6, Transformers 4.57.6, and vLLM 0.19.0 for the Qwen3.5 endpoint used in the VLM reproduction pipeline. The shared online memory is a two-layer GELU MLP ($\mathbb{R}^{768}\!\to\!\mathbb{R}^{768}$) updated only at test time; PLIP fusion uses the fixed default $\alpha{=}0.5$; main results enable the optional $k{=}3$ case archive over the $\sim$4\,MB in-domain index; no reflection round is used unless explicitly stated. Routing overrides and budget caps are listed in Appendix~\ref{sec:appendix:routing}.

\vspace{1mm}
\noindent\textbf{Metric details.}
WSI-VQA reports Total accuracy over all $735$ questions, MCQ accuracy over the $390$ multiple-choice questions, open-ended accuracy over the $345$ free-form questions, and compact text-generation scores. All WSI-VQA accuracy, BLEU, and ROUGE entries are reported as percentages. The main table keeps BLEU-1 and ROUGE-L, while Table~\ref{tab:appendix:wsivqa_full} also reports BLEU-4. BCNB uses letter-exact scoring for each diagnostic sub-task and the official weighted Overall score. PathMMU uses letter-exact validation accuracy. Efficiency metrics are per-question median latency, peak adjudicator-token count, peak GPU memory when tabulated, and offline footprint; lower is better.

\subsection{Baseline Settings, Model Names, and API Scope}
\label{sec:appendix:baseline_settings}
\label{sec:appendix:model_scope}

This subsection combines the reproducibility-facing baseline table with the exact closed-source and open-source model identifiers used by the main comparison.

\begin{table}[H]
\centering
\caption{Detailed settings behind the shortened method names
used in the main result tables. ``Published'' means that we did
not re-run the method under our local pipeline. All local
reproduced rows use Qwen3.5-4B as the common final
adjudicator / answer-normalization layer.}
\label{tab:appendix:settings}
\small
\setlength{\tabcolsep}{3.5pt}
\renewcommand{\arraystretch}{1.12}
\begin{tabularx}{\linewidth}{@{}>{\raggedright\arraybackslash}p{0.19\linewidth}>{\raggedright\arraybackslash}p{0.15\linewidth}>{\raggedright\arraybackslash}p{0.17\linewidth}X@{}}
\toprule
Name in main tables & Source & Visual input & Backbone / setting \\
\midrule
\multicolumn{4}{l}{\textit{Pretrained MLLM Models}} \\
GPT-4o & local API run & slide thumbnail & OpenAI endpoint \texttt{gpt-4o} \\
GPT-5.4 & local API run & slide thumbnail & OpenAI endpoint \texttt{gpt-5.4} \\
Gemini-3.1-Pro-preview & local API run & slide thumbnail & Google endpoint \texttt{gemini-3.1-pro-preview} \\
Qwen3.5 & reproduced & slide thumbnail & Qwen/Qwen3.5-4B and Qwen/Qwen3.5-9B \\
\midrule
\multicolumn{4}{l}{\textit{Supervised Pathology Models}} \\
WSI-VQA model & reproduced & WSI-level & official WSI-VQA model \\
SlideChat & reproduced & slide-level & trained slide MLLM \\
WSI-LLaVA & reproduced & patch or slide thumbnail & official setting / local patch input \\
MedDr~\citep{he2024gsco} & reproduced & slide thumbnail & Slide-T setting \\
TITAN~\citep{TITAN} & reproduced & slide thumbnail / CONCH features & zero-shot CONCH~v1.5~\citep{lu2024conch} features \\
Quilt-LLaVA~\citep{quiltllava2024} & reproduced & patch input & official checkpoint \\
LLaVA-Med~\citep{li2023llava} & reproduced & patch input & official checkpoint \\
MedGemma~\citep{sellergren2025medgemma} & reproduced & slide thumbnail & MedGemma 1.5-4B \\
Patho-R1 only & reproduced & selected patches & Patho-R1-7B~\citep{zhang2026pathor1} evidence + Qwen3.5-4B~\citep{qwen2026qwen35} adjudicator \\
\midrule
\multicolumn{4}{l}{\textit{Training-Free Pathology Agents}} \\
PathAgent & reproduced & patch trajectory & official pipeline + Patho-R1-7B~\citep{zhang2026pathor1} / Qwen3.5-4B~\citep{qwen2026qwen35} join \\
\name & reproduced & scan-search-readout & Patho-R1-7B~\citep{zhang2026pathor1} / Qwen3.5-4B~\citep{qwen2026qwen35}, Appendix~\ref{sec:appendix:routing} \\
\bottomrule
\end{tabularx}
\end{table}

Pretrained MLLM rows in the main tables include closed-source API endpoints and local Qwen thumbnail runs, all using the same single-thumbnail (Slide-T) visual input. The closed-source rows are identified by the exact endpoint labels used in our runs, while the citations point to the closest public model-family references available for each provider. Thus the citation records the family/version context rather than a separate reproducibility guarantee for every endpoint suffix. Specifically, we used \texttt{gpt-5.4}, \texttt{gemini-3.1-pro-preview}, \texttt{Qwen/Qwen3.5-4B}, and \texttt{Qwen/Qwen3.5-9B}. For MedGemma, the main tables keep MedGemma-1.5-4B as the size-comparable medical VLM baseline and omit the much larger 27B variant from the primary comparison.

\subsection{PathAgent Reproduction Protocol}
\label{sec:appendix:pathagent}

The PathAgent row in the main paper uses the official codebase configured with Patho-R1-7B~\citep{zhang2026pathor1}, Qwen3.5-4B~\citep{qwen2026qwen35}, and our evaluation join. This keeps the direct agent-to-agent comparison on equal footing rather than anchoring it to the original paper-reported MCQ-only number. We keep the detailed protocol note here, rather than in the introduction, so the main paper stays focused on \name itself.

\begin{table}[H]
\centering
\caption{Protocol note for the PathAgent row. The row is used
as a controlled agent-loop comparison under our evaluation stack, not as
an exact reproduction of every setting in the original PathAgent paper.}
\label{tab:appendix:pathagent_protocol}
\small
\setlength{\tabcolsep}{5pt}
\begin{tabular*}{\linewidth}{@{\extracolsep{\fill}}lp{0.68\linewidth}}
\toprule
Item & Setting in this paper \\
\midrule
Code path & Official PathAgent implementation configured locally. \\
Question set & Same WSI-VQA test split and question records as Table~\ref{tab:wsivqa}. \\
Answer stack & Same frozen Patho-R1-7B perceptor and Qwen3.5-4B adjudicator used for \name; all local reproduced rows use Qwen3.5-4B as the final answer-normalization layer. \\
Evaluation join & Same prediction join, answer parsing, SeqMatcher-based MCQ scoring, and BLEU/ROUGE scripts used for \name. \\
Reported use & Direct comparison to \name uses this controlled protocol. The original paper-reported MCQ value is cited only as context because it used a different release/evaluation setting. \\
\bottomrule
\end{tabular*}
\vspace{-0.5em}
\end{table}

\section{Method Details}
\label{sec:appendix:method_details}

After the dataset, baseline, and reproduction context, this section records the deferred method-level material: the question-aware routing and budget heuristics, the Navigation Summary prompt, and the full inference pseudocode.

\subsection{Question-aware Routing and Budget Allocation}
\label{sec:appendix:routing}

The main paper refers to the scan controller as a lightweight dynamic routing and budget-allocation mechanism. This controller does not change backbone weights or introduce question-specific model parameters. It only adjusts spatial coverage and evidence budget using a few pre-scan heuristics derived from the current question.

The released code exposes global parser defaults, but the paper reports the launch configuration used for the main experiments. Table~\ref{tab:appendix:routing_config} records the reported overrides in lookup form.

\begin{table}[H]
\centering
\caption{Reported routing, memory, and evidence-budget settings used by the main experiments.}
\label{tab:appendix:routing_config}
\footnotesize
\setlength{\tabcolsep}{4pt}
\begin{tabular}{p{0.22\linewidth}p{0.70\linewidth}}
\toprule
Component & Reported setting \\
\midrule
Question router & Keyword heuristic mapping each question to morphology, clinical, or other. \\
Scan-time coverage & NMS distance is $\max(d_{\min},4096)$ for morphology, $\max(d_{\min},20480)$ for clinical, and $\max(d_{\min},0.08\,\mathrm{diag}(\mathcal{W}))$ otherwise. \\
First-pass search budget & Base $K_0=10$; $K_{\mathrm{search}}(q)=K_0+2$ for morphology, $K_0+3$ for clinical, and $K_0$ otherwise. Pre-rerank pool $K_{\mathrm{pool}}=\max(30,K_{\mathrm{search}}(q)R_{\max})$ with $R_{\max}=1$. \\
Shared online memory & Reinitialized once per slide with PyTorch defaults; two-layer GELU MLP, hidden width $768$, learning rate $0.05$, gradient clipping $5.0$, forgetting factor $\alpha_f=0.999$, warm-up $T_w=100$, threshold scale $\lambda=1.0$. \\
Local high-mag memory & Reinitialized for each ROI with warm-up $\min(\max(T_w/5,5),|\mathcal{P}_{h}(r)|)$. \\
Evidence and adjudication budget & $T=2$ local patches per ROI, global perceptor cap $V_{\max}=15$, and one adjudication round. Reflection is disabled in main results; exploratory reflection runs split remaining targets evenly across remaining rounds. \\
\bottomrule
\end{tabular}
\vspace{-0.5em}
\end{table}

\subsection{Navigation Summary Prompt Example}
\label{sec:appendix:nav_prompt}

The Navigation Summary is a short prompt field inserted into the adjudicator input. It is not a learned memory module and does not expose the full scan trace. We separate the reusable prompt slot from the filled qualitative example so the reader can distinguish format from case-specific evidence.

\noindent\textbf{Prompt slot template.}
\begin{quote}
\small
\textbf{Navigation Summary.} Low-magnification scan statistics: mean \texttt{<mean>}, standard deviation \texttt{<std>}, high-surprise fraction \texttt{<fraction>}, and \texttt{<count>} candidate regions after thresholding and NMS. Use this as slide-level context, but base the final answer on the regional evidence below.
\end{quote}

\noindent\textbf{Filled qualitative example.}
For the nuclear-grade case in Figure~\ref{fig:pathagent_case_compare}, the adjudicator receives the Navigation Summary together with selected regional evidence:

\begin{quote}
\small
\textbf{Question.} What is the nuclear grade of the invasive ductal carcinoma?\\
\textbf{Navigation Summary.} The scan produced a compact set of high-surprise candidates after distance-based NMS. The selected targets repeatedly concentrate on tumor-rich high-magnification regions rather than coarse local fields.\\
\textbf{Regional Evidence.} Evidence patches describe irregular nests, moderate nuclear pleomorphism, hyperchromasia, sparse mitoses, and partial tubule loss.\\
\textbf{Reference Context.} Optional retrieved cases use similar wording for moderate nuclear grade.\\
\textbf{Adjudication instruction.} Answer the question directly and use the evidence above to choose the most compatible label.
\end{quote}

\subsection{Inference Pseudocode}
\label{sec:appendix:method_algorithm}

\begin{algorithm}[H]
\caption{Training-free inference in \name}
\label{alg:appendix:inference}
\small
\begin{algorithmic}[1]
\State \textbf{Input:} slide $\mathcal{W}$, question $q$, frozen feature extractors and frozen answer stack
\State Extract low-magnification tiles $\mathcal{X}_{\ell}(\mathcal{W})$ and initialize fresh online memory $\mathcal{M}_{\boldsymbol{\theta}_0}$
\State Scan low-magnification tiles; before each update, compute surprise $\sigma_i$ from the reconstruction-gradient norm
\State After warm-up, update memory only for tiles with $\sigma_i>\tau_{\mathrm{sur}}$ and record the surprise history
\State Apply thresholding and distance-based NMS with spacing $\rho(q)$ to obtain ROI pool $\mathcal{R}(q)$
\State Compute PLIP relevance inside $\mathcal{R}(q)$ and combine it with surprise to rank candidate regions
\State Select $K_{\mathrm{search}}(q)$ targets as $\mathcal{C}(q)$
\State For each selected target, initialize a local high-magnification memory and keep the top-$T$ local surprise patches
\State Aggregate selected patches into evidence set $\mathcal{E}(q)$ and construct the Navigation Summary from scan statistics
\State Optionally retrieve Reference Context from the in-domain case archive
\State Use the perceptor to convert selected patches into regional evidence strings
\State Use the adjudicator to answer from regional evidence, Navigation Summary, and optional Reference Context
\State \textbf{Output:} final answer $a$
\end{algorithmic}
\end{algorithm}

\section{Experimental Results and Additional Analyses}
\label{sec:appendix:experimental_results}

This section groups all supplementary evidence after the experimental setup. We place the complete numeric tables first, then provide compact visual summaries, workflow illustrations, ablations, efficiency details, and negative findings.

\subsection{Full Result Tables}
\label{sec:appendix:tables}

\begin{table}[H]
\centering
\caption{Full WSI-VQA table including BLEU-4. All numeric scores are percentages. TITAN reports
only MCQ because this zero-shot setting can only be mapped to
answer options.}
\label{tab:appendix:wsivqa_full}
\small
\setlength{\tabcolsep}{5pt}
\resizebox{\linewidth}{!}{%
\begin{tabular}{lcccccc}
\toprule
Method & Total & MCQ & Open & BLEU-1 & BLEU-4 & ROUGE-L \\
\midrule
\multicolumn{7}{l}{\textit{Pretrained MLLM Models}} \\
GPT-4o~\citep{openai2024gpt4o}                & $27.57$ & $32.50$ & $22.00$ & $8.53$ & $3.84$ & $23.62$ \\
GPT-5.4~\citep{openai2026models}              & $35.74$ & $42.18$ & $28.46$ & $14.27$ & $7.13$ & $28.86$ \\
Gemini-3.1-Pro-preview~\citep{google2026gemini31pro} & $34.41$ & $40.42$ & $27.62$ & $12.84$ & $6.42$ & $27.56$ \\
Qwen3.5-4B~\citep{qwen2026qwen35} (Slide-T) & $40.26$ & $40.26$ & $40.26$ & $9.21$ & $4.53$ & $24.13$ \\
Qwen3.5-9B~\citep{qwen2026qwen35} (Slide-T) & $38.50$ & $49.23$ & $26.38$ & $57.31$ & $57.08$ & $48.04$ \\
\midrule
\multicolumn{7}{l}{\textit{Supervised Pathology Models}} \\
WSI-VQA model~\citep{chen2024wsi}             & $42.18$ & $46.90$ & $36.84$ & $10.83$ & $5.21$ & $31.79$ \\
MedDr~\citep{he2024gsco} (Slide-T)           & $45.18$ & $43.69$ & $46.86$ & $36.82$ & $17.83$ & $45.24$ \\
TITAN~\citep{TITAN} (zero-shot, CONCH v1.5 feats; MCQ-only) & ---     & $34.87$ & ---     & ---     & ---     & ---     \\
Quilt-LLaVA~\citep{quiltllava2024} (Patch)   & $30.07$ & $29.23$ & $31.01$ & $0.24$ & $0.12$ & $25.13$ \\
LLaVA-Med~\citep{li2023llava} (Patch)        & $21.22$ & $35.64$ & $4.93$  & $0.31$ & $0.11$ & $22.74$ \\
WSI-LLaVA~\citep{liang2025wsi} (Patch)      & $33.20$ & $42.82$ & $22.32$ & $5.73$ & $2.71$ & $31.24$ \\
MedGemma 1.5-4B~\citep{sellergren2025medgemma} (Slide-T) & $42.99$ & $53.08$ & $31.59$ & $18.04$ & $10.12$ & $34.93$ \\
\midrule
\multicolumn{7}{l}{\textit{Training-Free Pathology Agents}} \\
PathAgent~\citep{chen2025pathagent} & $33.88$ & $51.54$ & $13.91$ & $42.73$ & $38.91$ & $33.54$ \\
\textbf{\name (ours)}                     & \boldmath$56.34$ & \boldmath$52.21$ & \boldmath$61.00$ & \boldmath$60.92$ & \boldmath$50.74$ & \boldmath$63.53$ \\
\bottomrule
\end{tabular}
}
\end{table}

\begin{table}[H]
\centering
\caption{Full SlideBench-BCNB breakdown with merged HER2 and delta rows.
HER2 averages the HER2-status and HER2-expression sub-tasks as in
Table~\ref{tab:bcnb}; Overall keeps the official seven-task weighting.
All deltas are absolute percentage points relative to Patho-R1 alone.}
\label{tab:appendix:bcnb_full}
\small
\setlength{\tabcolsep}{3pt}
\resizebox{\linewidth}{!}{%
\begin{tabular}{lccccccc}
\toprule
Method & Overall & Tumor & ER & PR & HER2 & Grading & Mol.Sub \\
\midrule
\multicolumn{8}{l}{\textit{Pretrained MLLM Models}} \\
GPT-4o~\citep{openai2024gpt4o}      & $43.78$ & $56.84$ & $49.96$ & $49.42$ & $42.13$ & $38.46$ & $26.84$ \\
GPT-5.4~\citep{openai2026models}    & $45.85$ & $60.42$ & $50.42$ & $49.86$ & $44.51$ & $42.36$ & $28.43$ \\
Gemini-3.1-Pro-preview~\citep{google2026gemini31pro} & $44.91$ & $58.46$ & $50.18$ & $49.62$ & $43.58$ & $40.85$ & $27.62$ \\
Qwen3.5-4B~\citep{qwen2026qwen35}            & $39.48$ & $49.43$ & $33.36$ & $27.32$ & \runner{48.02} & $52.70$ & $19.19$ \\
Qwen3.5-9B~\citep{qwen2026qwen35}            & $41.99$ & $53.84$ & $36.42$ & $30.18$ & $47.53$ & \best{59.15} & $21.43$ \\
\midrule
\multicolumn{8}{l}{\textit{Supervised Pathology Models}} \\
SlideChat~\citep{chen2025slidechat} & $54.07$ & \best{89.90} & \runner{72.98} & \best{72.98} & $47.53$ & $23.11$ & $20.58$ \\
WSI-LLaVA~\citep{liang2025wsi}      & $52.68$ & $85.32$ & $60.81$ & $60.81$ & $42.78$ & $46.28$ & $29.20$ \\
MedDr~\citep{he2024gsco}            & $35.48$ & $59.60$ & $43.80$ & $39.20$ & $27.40$ & $29.40$ & $20.80$ \\
TITAN~\citep{TITAN} (zero-shot, CONCH v1.5 feats)          & $44.39$ & $85.16$ & $42.44$ & $41.68$ & $36.86$ & $45.35$ & $22.49$ \\
Quilt-LLaVA~\citep{quiltllava2024}           & $34.11$ & $33.46$ & $49.43$ & $44.05$ & $25.85$ & $34.99$ & $25.24$ \\
MedGemma 1.5-4B~\citep{sellergren2025medgemma} & $50.38$ & $79.68$ & $52.36$ & $60.35$ & $47.85$ & $42.33$ & $21.27$ \\
Patho-R1 alone (\name's perceptor, no agent) & $37.26$ & $42.16$ & $54.63$ & $50.38$ & $25.62$ & $35.53$ & $26.65$ \\
\midrule
\multicolumn{8}{l}{\textit{Training-Free Pathology Agents}} \\
PathAgent~\citep{chen2025pathagent}          & \runner{55.72} & $87.52$ & $63.69$ & $63.69$ & $44.52$ & \runner{55.95} & \runner{30.21} \\
\textbf{\name (ours)}            & \best{59.42} & \runner{88.28} & \best{76.09} & \runner{69.00} & \best{48.40} & $52.58$ & \best{32.33} \\
$\Delta$ vs Patho-R1 alone (pp)  & \boldmath$+22.16$ & \boldmath$+46.12$ & \boldmath$+21.46$ & \boldmath$+18.62$ & \boldmath$+22.78$ & $+17.05$ & \boldmath$+5.68$ \\
\bottomrule
\end{tabular}
}
\end{table}

\begin{table}[H]
\centering
\caption{Patch-level pathology MCQ transfer on PathMMU. Reported as a
single accuracy column with letter-exact scoring. The \name row is the
final local available-set run; most other rows are local available-set
runs, while the Quilt-LLaVA row is taken from the PathMMU paper. We
restrict the comparison to open-source pathology / general-purpose VLMs
or paper-reported open-source baselines; closed-source frontier APIs are
omitted because their PathMMU numbers are not reproducible under our
protocol.}
\label{tab:appendix:pathmmu_full}
\small
\setlength{\tabcolsep}{8pt}
\begin{tabular*}{\linewidth}{@{\extracolsep{\fill}}lc}
\toprule
Method & PathMMU val Acc \\
\midrule
Random choice                                                & $25.00$ \\
\midrule
\multicolumn{2}{l}{\textit{Pretrained MLLM Models}} \\
Qwen3.5-4B~\citep{qwen2026qwen35}                             & $38.16$ \\
Qwen3.5-9B~\citep{qwen2026qwen35}                             & $45.84$ \\
\midrule
\multicolumn{2}{l}{\textit{Supervised Pathology Models}} \\
Quilt-LLaVA~\citep{quiltllava2024}                           & $26.50$ \\
TITAN~\citep{TITAN}                                         & $32.18$ \\
WSI-LLaVA~\citep{liang2025wsi}                               & $36.18$ \\
SlideChat~\citep{chen2025slidechat}                          & $37.42$ \\
Patho-R1 only~\citep{zhang2026pathor1}                       & $42.18$ \\
MedGemma-1.5-4B~\citep{sellergren2025medgemma}               & $50.62$ \\
\midrule
\multicolumn{2}{l}{\textit{Training-Free Pathology Agents}} \\
PathAgent~\citep{chen2025pathagent}                          & $52.84$ \\
\textbf{\name (ours)}                                        & \boldmath$53.60$ \\
\bottomrule
\end{tabular*}
\vspace{-0.5em}
\end{table}

\begin{table}[H]
\centering
\caption{Full numeric $\alpha$ sweep on WSI-VQA. All scores are percentages. The main
paper summarizes the endpoint and default rows in
Table~\ref{tab:ablation_wsivqa}. The $\alpha{=}0$ row is
the no-PLIP setting because the ranking is driven only by
the surprise prior, and $\alpha{=}1$ is the PLIP-only
endpoint.}
\label{tab:appendix:alpha_numeric}
\small
\setlength{\tabcolsep}{6pt}
\begin{tabular*}{\linewidth}{@{\extracolsep{\fill}}lcccc}
\toprule
$\alpha$ & Total & MCQ & Open & BLEU-1 \\
\midrule
$0.00$ (surprise-only / no PLIP) & $53.42$ & $47.20$ & $60.46$ & $58.72$ \\
$0.25$ & $54.21$ & $49.42$ & $59.62$ & $59.43$ \\
\textbf{$0.50$} \emph{(default, \name)} & \boldmath$56.34$ & \boldmath$52.21$ & \boldmath$61.00$ & \boldmath$60.92$ \\
$0.75$ & $55.05$ & $50.50$ & $60.20$ & $59.83$ \\
$1.00$ (PLIP-only) & $52.20$ & $46.90$ & $58.20$ & $58.04$ \\
\bottomrule
\end{tabular*}
\vspace{-0.5em}
\end{table}

\begin{table}[H]
\centering
\caption{Candidate-pool constructor ablation on BCNB. Full
uses Table~\ref{tab:bcnb}; Receptor is avg(ER, PR); other rows
replace only candidate-pool construction.}
\label{tab:appendix:pool_constructor_bcnb}
\small
\setlength{\tabcolsep}{4pt}
\resizebox{\linewidth}{!}{%
\begin{tabular}{lccccc}
\toprule
Pool constructor & Total & Tumor & Receptor & Grading & Mol.Sub \\
\midrule
Full \name (surprise pool) & \best{59.42} & \best{88.28} & \best{72.55} & \best{52.58} & \best{32.33} \\
PLIP-built                 & \runner{57.61} & \runner{88.00} & \runner{70.90} & \runner{44.60} & \runner{29.60} \\
random                     & $54.16$ & $87.60$ & $68.80$ & $40.20$ & $23.20$ \\
\midrule
$\Delta$ PLIP $-$ Full \name   & $-1.81$ & $-0.28$ & $-1.65$ & $-7.98$ & $-2.73$ \\
$\Delta$ random $-$ Full \name & $-5.26$ & $-0.68$ & $-3.75$ & $-12.38$ & $-9.13$ \\
\bottomrule
\end{tabular}
}
\vspace{-0.5em}
\end{table}

\begin{table}[H]
\centering
\caption{Memory architecture comparison. This table is
deferred from the main paper because the main claim is
structural; the appendix keeps the full footprint details.}
\label{tab:appendix:memory}
\small
\setlength{\tabcolsep}{4pt}
\resizebox{\linewidth}{!}{%
\begin{tabular}{lccc}
\toprule
System & Offline artefacts & Runtime substrate & Tokens/Q \\
\midrule
PathAgent (training-free, query-driven; reported)    & $277$\,MB caption cache  & per-Q trajectory   & $\sim\!30$K \\
Patho-AgenticRAG (RL-enhanced agentic RAG; reported) & multi-GB ColQwen2 KB     & router + KB rows   & not reported \\
TissueLab (co-evolving tool agent; reported)         & curated tool factories   & HDF5 intermediates & not reported \\
SlideChat (trained slide MLLM; reported)             & 176K-VQA training corpus & single forward     & not comparable \\
\textbf{\name (training-free, ours; measured)}    & \textbf{4\,MB in-domain case archive} & \textbf{$<\!2$\,MB online-memory MLP} & \textbf{$\sim\!17$K} \\
\bottomrule
\end{tabular}
}
\end{table}

SlideChat is listed for offline-footprint context, but its
token accounting is not directly comparable: it is a trained
single-forward slide MLLM rather than a training-free agent
with per-question navigation, evidence selection, and
adjudication calls.

\subsection{Supplementary Comparison Plots}
\label{sec:appendix:comparison_plots}

The main paper reports full numeric tables and the
same-slide qualitative PathAgent comparison. Here we keep
compact visual summaries that are useful for quick scanning
after the full result tables.

\begin{figure}[H]
\centering
\includegraphics[width=0.92\linewidth]{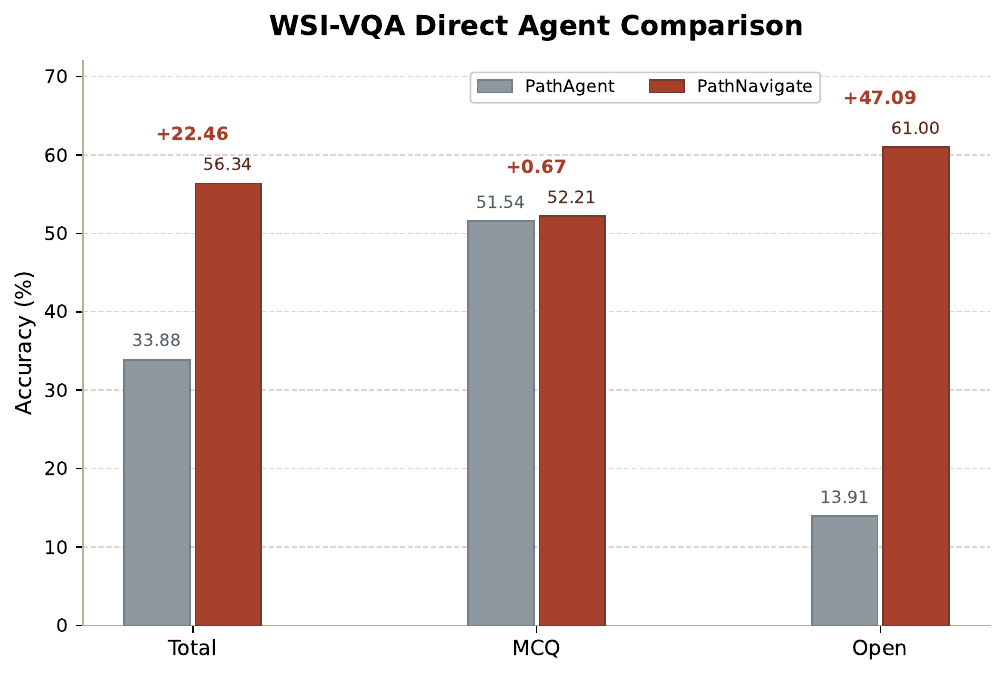}
\caption{Agent-level WSI-VQA comparison between the PathAgent row and \name under the same Patho-R1/Qwen3.5 evaluation join. Bars report Total, MCQ, and open-ended accuracy; the largest separation is on open-ended answering, while MCQ remains relatively close.}
\label{fig:appendix:wsivqa_agent_compare}
\end{figure}

\begin{figure}[H]
\centering
\includegraphics[width=0.96\linewidth]{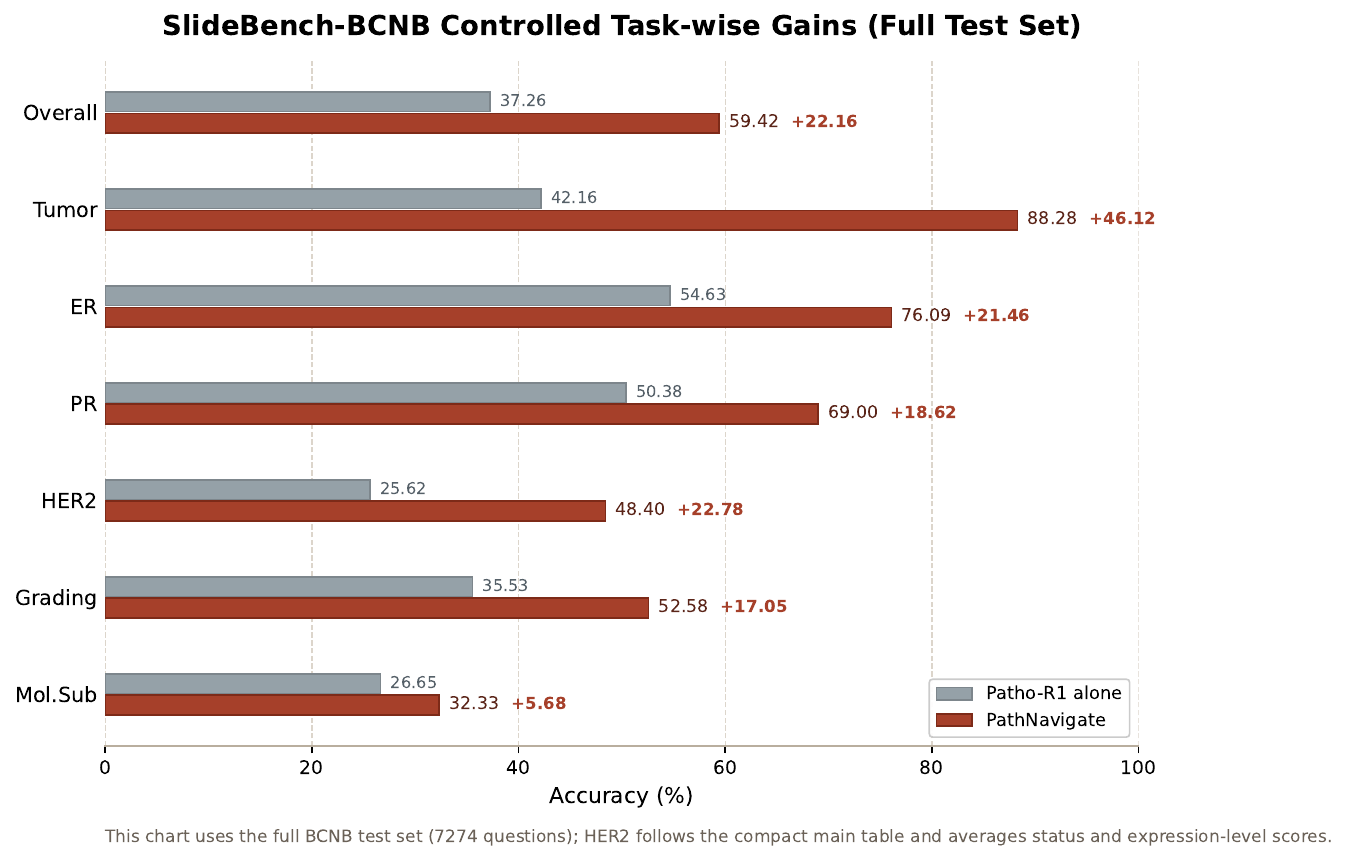}
\caption{Task-wise SlideBench-BCNB comparison between Patho-R1 alone and \name under the unified protocol. The largest gains appear on Tumor, ER, PR, and the merged HER2 score, matching the compact main-table reporting.}
\label{fig:appendix:bcnb_taskwise}
\end{figure}

\subsection{Workflow Gallery Template}
\label{sec:appendix:workflow}

The supplementary qualitative gallery uses a fixed panel
layout so that cases can be compared on equal visual
footing. Each case includes seven blocks in a constant
order: slide overview, surprise prior, PLIP relevance,
combined ranking, top low-magnification candidates, selected
evidence patches, and the final answer summary. The main
paper uses the same-slide comparison in
Figure~\ref{fig:pathagent_case_compare}; the single-system
workflow panel below shows the full internal trace for
\name.

\begin{figure*}[t]
\centering
\includegraphics[width=0.98\textwidth]{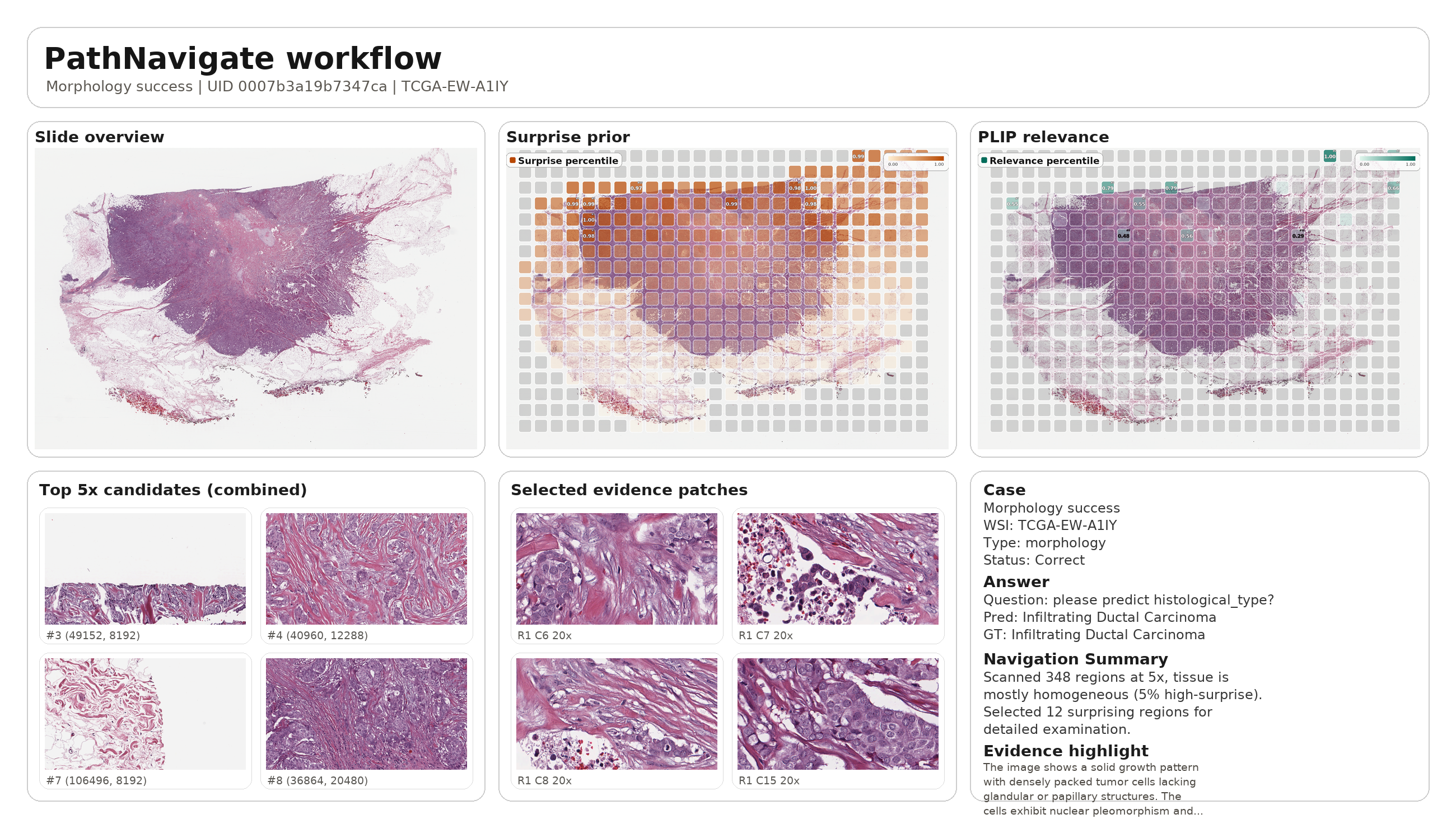}
\caption{Representative workflow case for \name on WSI-VQA.
From left to right, the panel shows the slide overview,
surprise prior, PLIP relevance, combined within-pool
ranking, top low-magnification candidate regions, selected evidence
patches, and the final answer summary. Heatmaps are normalized within
the slide, so stronger saturation denotes higher surprise or relevance
within that case.}
\label{fig:appendix:workflow_case}
\end{figure*}

The fixed layout is retained as a gallery template: additional exported
cases use the same seven-panel semantics rather than hand-curated
screenshots with case-specific ordering.

\subsection{Sensitivity and Component Figures}
\label{sec:appendix:sensitivity_figures}

We group the remaining claim-bearing plots here so the main
paper can stay table-led while the appendix still exposes
the main shape of the ablation evidence. The corresponding
numeric rows are collected in Tables~\ref{tab:appendix:alpha_numeric},
\ref{tab:appendix:pool_constructor_bcnb}, \ref{tab:appendix:highmag_readout}, and~\ref{tab:appendix:mlp_arch},
with negative-result checks in Appendix~\ref{sec:appendix:negative}.

\begin{figure}[H]
\centering
\includegraphics[width=0.95\linewidth]{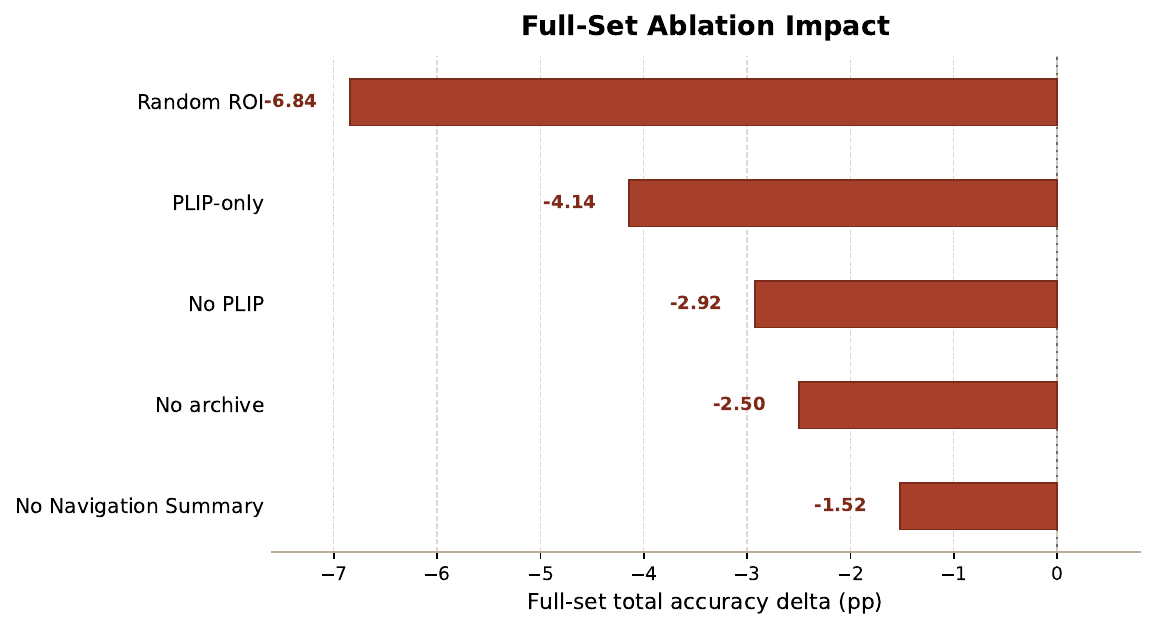}
\caption{Compact full-set ablation impact summary. Each bar
shows the total-accuracy delta against the full-set
\name reference for that ablation. The archive-off and
Navigation-Summary-off rows quantify the answer-time context
paths, while the PLIP endpoint and no-PLIP setting show that
question-conditioned reranking is useful but not a standalone
replacement for the scan prior.}
\label{fig:appendix:ablation_summary}
\end{figure}

\begin{figure}[H]
\centering
\includegraphics[width=0.95\linewidth]{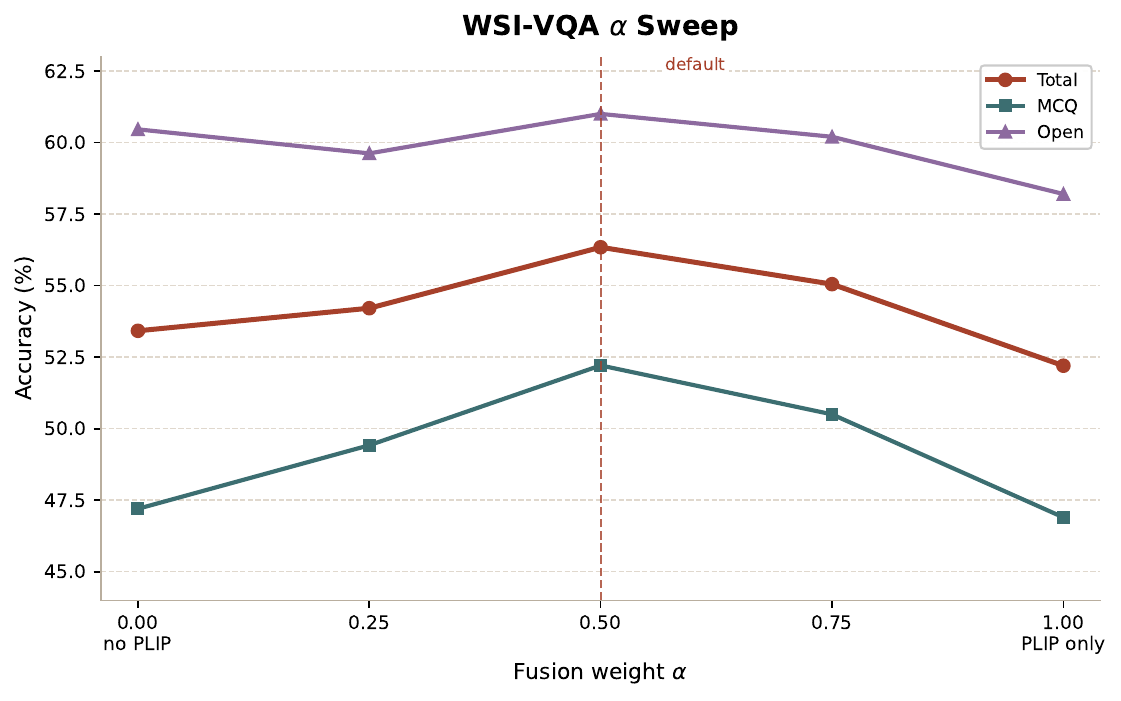}
\caption{$\alpha$ sweep on WSI-VQA. The default
$\alpha{=}0.5$ lies on the empirical peak, while the
endpoint drops support the interpretation that PLIP is most
useful as a within-pool refinement signal rather than as the
primary proposal rule.}
\label{fig:appendix:alpha_sweep}
\end{figure}

\begin{table}[H]
\centering
\caption{High-magnification evidence readout ablation. The
same selected low-magnification targets are held fixed while
the evidence readout is changed. BCNB reports the aggregate
Overall score.}
\label{tab:appendix:highmag_readout}
\small
\setlength{\tabcolsep}{6pt}
\begin{tabular*}{0.78\linewidth}{@{\extracolsep{\fill}}lcccc}
\toprule
Readout & Total & MCQ & Open & BCNB \\
\midrule
Full & \best{56.34} & \best{52.21} & \best{61.00} & \best{59.42} \\
Global mem. & $53.32$ & $48.86$ & $58.36$ & $56.33$ \\
Random local & \runner{54.12} & \runner{49.62} & \runner{59.20} & \runner{57.50} \\
Low-mag only & $51.01$ & $45.86$ & $56.84$ & $52.37$ \\
\bottomrule
\end{tabular*}
\vspace{-0.5em}
\end{table}

This ablation isolates readout after target selection. Fresh
local high-magnification memory outperforms global reuse,
random local sampling, and low-magnification-only evidence,
showing that selected ROIs still need nucleus-resolution
morphology.

\begin{table}[H]
\centering
\caption{Shared online-memory MLP architecture sensitivity on WSI-VQA.
All five variants were rerun under the same sweep
protocol and each reports $726$ answered questions with
$9$ skips, so the numbers are read as a within-sweep
comparison rather than mixed directly with the main-result
row. The default two-layer equal-width MLP remains best.}
\label{tab:appendix:mlp_arch}
\small
\setlength{\tabcolsep}{5pt}
\begin{tabular*}{\linewidth}{@{\extracolsep{\fill}}lcccc}
\toprule
Variant & Total & Morphology & Clinical & Other \\
\midrule
L1H768            & $50.96$          & $58.40$          & $49.79$          & $45.10$          \\
L2H384            & $52.62$          & $61.76$          & $51.93$          & $44.71$          \\
L2H768 (default)  & $\mathbf{56.06}$ & $\mathbf{65.97}$ & $\mathbf{54.08}$ & $\mathbf{48.63}$ \\
L2H1536           & $53.44$          & $61.76$          & $51.93$          & $47.06$          \\
L3H768            & $54.68$          & $65.55$          & $50.64$          & $48.24$          \\
\bottomrule
\end{tabular*}
\vspace{-0.5em}
\end{table}

The sweep shows a simple pattern: collapsing the substrate
to a single linear layer hurts most, while adding width or
one extra layer does not improve over the default. This
supports the main-paper choice of a shallow but
non-degenerate two-layer online-memory MLP.

\begin{figure}[H]
\centering
\includegraphics[width=0.98\linewidth]{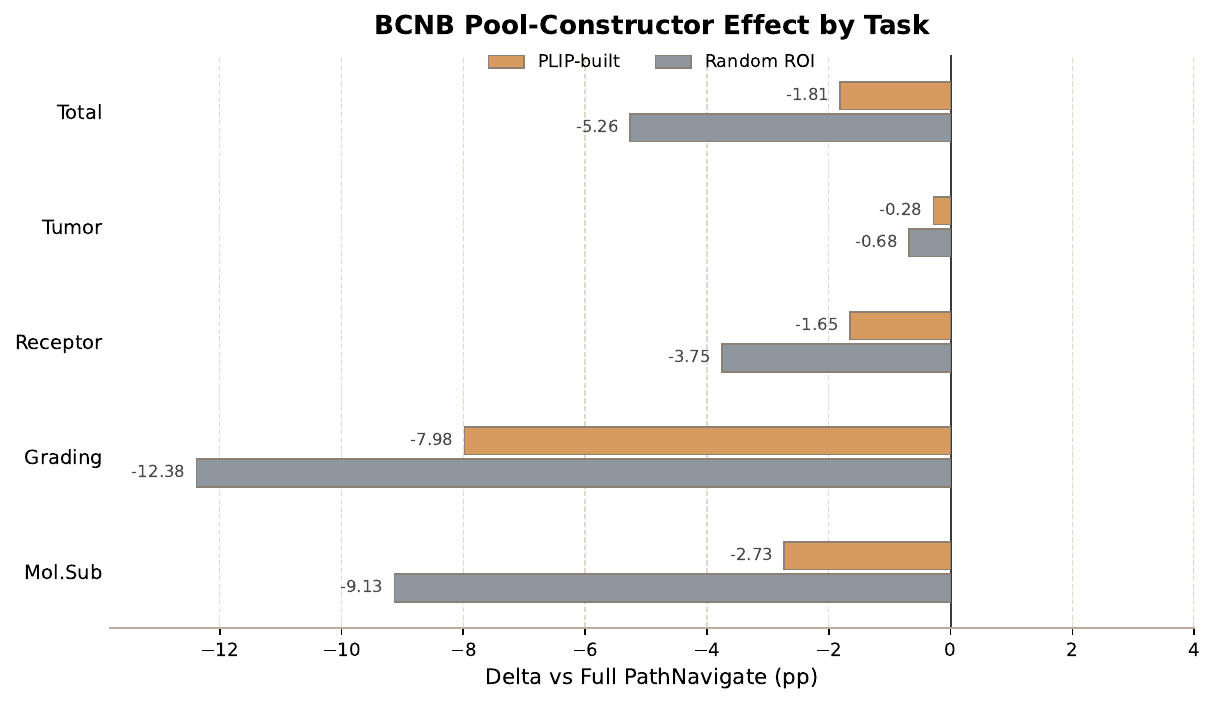}
\caption{Candidate-pool constructor effect on BCNB. Bars
show task-wise deltas relative to the Full \name main-result
row, making receptor and grading-sensitive changes visible without
over-emphasizing tumor-saturated categories where all constructors
are close.}
\label{fig:appendix:pool_constructor}
\end{figure}

\subsection{Efficiency Visualization}
\label{sec:appendix:efficiency_plots}

The main paper reports the training-free agent cost comparison as the bar chart in Figure~\ref{fig:efficiency}. This appendix keeps the internal breakdown so the archive and PLIP costs remain visible without expanding the main text.

\begin{table}[H]
\centering
\caption{Internal \name efficiency breakdown on WSI-VQA under
the same hardware and evaluation stack as Figure~\ref{fig:efficiency}.
These rows validate where the runtime and token costs enter,
but they are not separate systems in the main comparison.}
\label{tab:appendix:efficiency_internal}
\small
\setlength{\tabcolsep}{6pt}
\begin{tabular*}{\linewidth}{@{\extracolsep{\fill}}lccc}
\toprule
System & Latency (s/Q, median) & Peak GPU mem & Tokens/Q (peak) \\
\midrule
\name (reported default) & $\sim\!84$ & $\sim\!22$\,GB & $\sim\!17$K \\
\name $-$ archive (\texttt{abl-no-rag}) & $\sim\!28$ & $\sim\!22$\,GB & $\sim\!11$K \\
\name $-$ PLIP (\texttt{abl-no-plip}) & $\sim\!74$ & $\sim\!21$\,GB & $\sim\!17$K \\
\bottomrule
\end{tabular*}
\vspace{-0.5em}
\end{table}

\begin{figure}[H]
\centering
\includegraphics[width=0.98\linewidth]{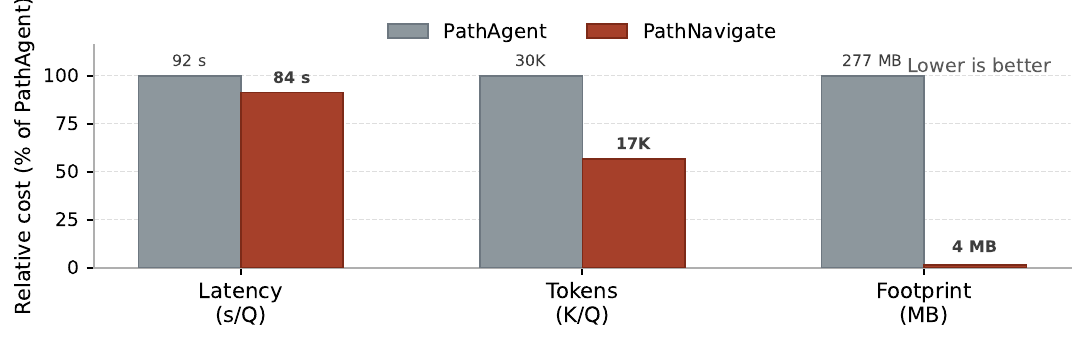}
\caption{Detailed per-question cost profile for training-free
agents, complementing the main-paper comparison in Figure~\ref{fig:efficiency}.
The breakdown keeps the direct PathAgent--\name comparison visible:
latency is similar, token use is lower, and the offline footprint is much smaller.}
\label{fig:appendix:efficiency_profile}
\end{figure}

\subsection{Negative Findings}
\label{sec:appendix:negative}

Two negative findings are worth recording even though they do not belong in the main story. We also fixed two implementation defaults from these checks before reporting the main experiments: a single adjudication pass matched or exceeded multi-round variants, and free-text perceptor evidence matched or exceeded a strict JSON schema.
\begin{itemize}[leftmargin=16pt,itemsep=2pt]
\item \textbf{Norm-preserving consolidation.} We tested an inverted update rule that only consolidated low-surprise patches. The idea was to protect a stable notion of normal tissue on tumor-heavy slides. In our setting, the change did not produce a reliable accuracy gain, so we keep it as a documented negative result rather than a main contribution.
\item \textbf{Episodic trace memory.} We also tested an extra memory layer that prepended the full per-step trace to the adjudicator prompt. This consistently made the prompts longer and did not help final accuracy. We therefore removed it from the reported system.
\end{itemize}

\end{document}